%% file: main.tex
\pdfoutput=1

\documentclass[11pt]{article}

\usepackage[]{acl}

\usepackage{times}
\usepackage{latexsym}

\usepackage[T1]{fontenc}

\usepackage[utf8]{inputenc}

\usepackage{microtype}

\usepackage{graphicx} 
\usepackage{multirow} 
\usepackage{amsmath}%
\usepackage{MnSymbol}%
\usepackage{wasysym}%
\usepackage{tikz}
\def\checkmark{\tikz\fill[scale=0.4](0,.35) -- (.25,0) -- (1,.7) -- (.25,.15) -- cycle;}
\usepackage{enumerate} 
\usepackage{paralist}
\usepackage{inconsolata}
\usepackage{booktabs}

\title{Desiderata for the Context Use of Question Answering Systems}

\author{Sagi Shaier,$^\diamond$ Lawrence E Hunter,$^\dag$ Katharina von der Wense$^{\nabla\diamondsuit}$ \\
  $^\nabla$University of Colorado Boulder\\
$^\dag$Independent Scholar \\
$^\diamondsuit$Johannes Gutenberg University Mainz\\
$^\nabla$E-mail: \{sagi.shaier, katharina.kann\}@colorado.edu \\
$^\dag$E-mail: Prof.Larry.Hunter@gmail.com \\
 \\}

\begin{document}
\maketitle
\begin{abstract}
Prior work has uncovered a set of common problems in state-of-the-art context-based question answering (QA) systems: a lack of attention to the context when the latter conflicts with a model's parametric knowledge, little robustness to noise, and a lack of consistency with their answers. However, most prior work focus on one or two of those problems in isolation, which makes it difficult to see trends across them. We aim to close this gap, by first outlining a set of -- previously discussed as well as novel -- desiderata for QA models. We then survey relevant analysis and methods papers to provide an overview of the state of the field. The second part of our work presents  experiments where we evaluate 15 QA systems on 5 datasets according to all desiderata \textit{at once}. We find many novel trends, including (1) systems that are less susceptible to noise are not necessarily more consistent with their answers when given irrelevant context; (2) most systems that are more susceptible to noise are more likely to correctly answer according to a context that conflicts with their parametric knowledge; and (3) the combination of conflicting knowledge and noise can reduce system performance by up to $96\%$. As such, our desiderata help increase our understanding of how these models work and reveal potential avenues for improvements. Code and data can be found here: \url{https://github.com/Shaier/context_usage_desiderata.git}.

\end{abstract}

\section{Introduction}
\begin{figure}[t]
    \includegraphics[width=1\columnwidth,height=0.7\columnwidth,keepaspectratio]{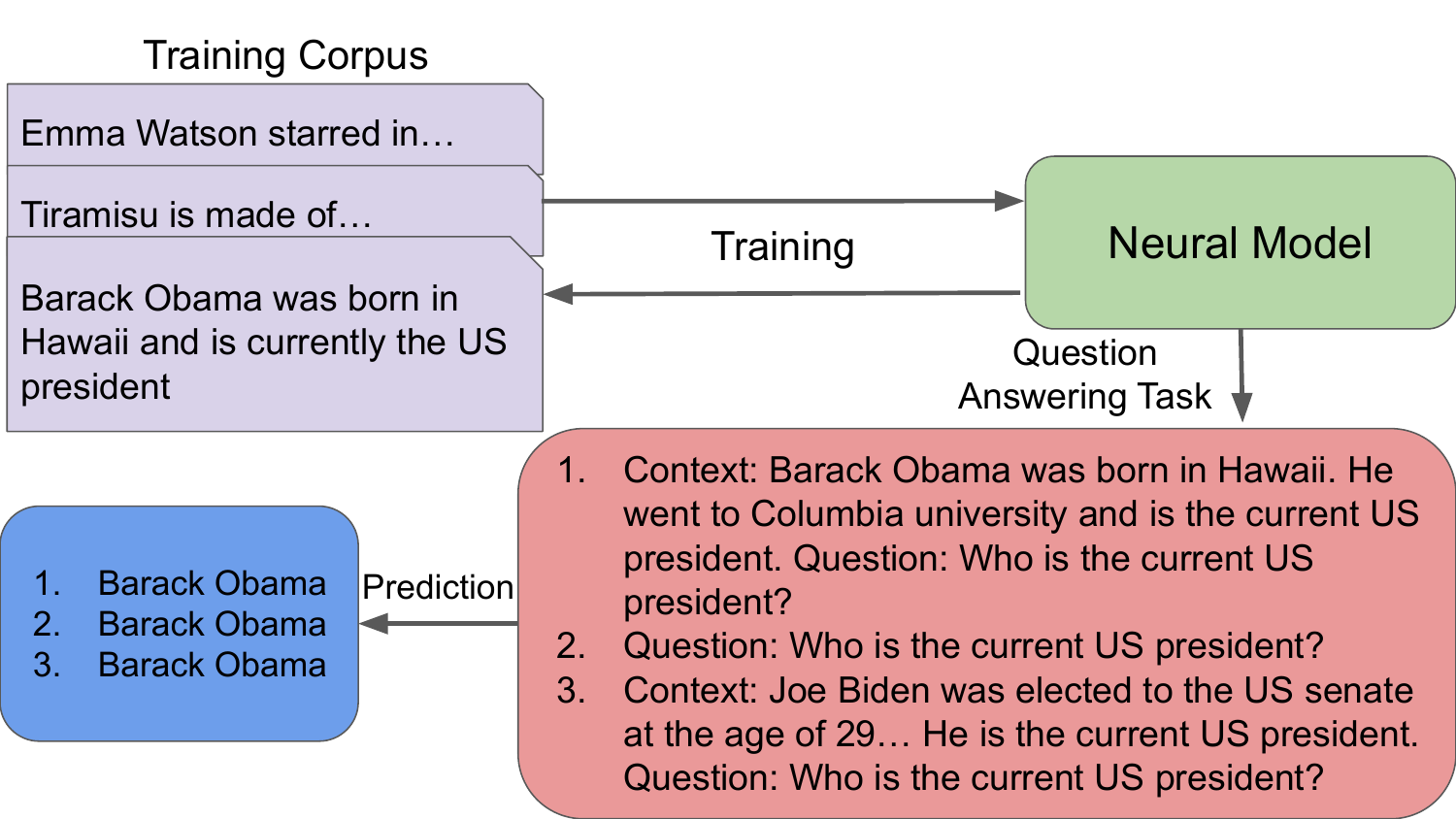}
    \caption{An example where the model was trained to learn the knowledge "Barack Obama is the current US president". In the first and second tasks the model answered the questions correctly. However, in the third task where the model is given context with conflicting information it fails to answer the question correctly.}
    \label{main_figure}
\end{figure}

Question answering (QA) systems which are based on large language models (LLMs) play a larger role than ever before in our society, due to their ability to offer quick access to information \cite{petroni, roberts-etal-2020-much, shin-etal-2020-autoprompt, sung-etal-2021-language, jiang2020know}. 
Many QA systems can make use of context information when available, which often contains relevant information to help systems answer questions, cf. Figure \ref{main_figure}.
We refer to all systems that are able to leverage such context information as \textit{context-based QA systems}.


Many aspects of such systems have been evaluated by previous work, such as the amount of their parametric knowledge \cite{petroni} and their robustness to noise \cite{jia-liang-2017-adversarial}, conflicting knowledge \cite{contra, shaier2024adaptivequestionansweringenhancing, longpre-etal-2021-entity}, or irrelevant contexts \cite{control, dist} . However, looking at such aspects in isolation makes it difficult to see trends across problems, e.g., to explore whether there is a connection between a model's attention to context and its ability to handle noise. 


Here, we 1) outline a set of -- previously discussed as well as novel -- desiderata for context-based QA models and 2) provide an extensive survey of related works, which we group and discuss according to our desiderata. Such desiderata unify some of the existing aspects from the literature, e.g., robustness to conflicting knowledge, 
and outline how a QA system should behave from the perspective of the context.
We will publicly release a toolkit to prepare datasets — both free-form and multiple choice (MC) type -- to evaluate models according to all desiderata \textit{at once}. 

Using our toolkit, we 3) evaluate 15 LLM-based QA systems and first confirm prior works' results: while some systems appear nearly perfect, scoring $99\%$ accuracy on standard datasets, their performance is significantly worse according to many of our desiderata. For instance, their accuracy drops by up to $93\%$ with noise, such as random strings as distractors. Second, considering all desiderata \textit{at once}, we find that (1) systems that are less susceptible to noise are more consistent with their answers when provided with irrelevant context; (2) most systems that are more susceptible to noise are more likely to correctly answer according to a context that conflicts with their parametric knowledge; and (3) the combination of conflicting knowledge and noise can reduce system performance by up to $96\%$. Finding these novel trends using our desiderata opens new avenues to improve QA models.

\section{Desiderata}
\label{ideal_QA}
\input{new_table1.tex}
We now 
develop a set of desiderata regarding the context use of a model, before presenting our survey on what prior work has found with regards to our desiderata and performing our own experiments in the next two sections. To come up with our desiderata, which are presented in Table \ref{desiderata_table}, we consider the question: \textit{How would an ideal QA model behave for different types of context?} 

The ideal behavior depends on whether the knowledge in the context is known or unknown to the model. For example, looking at Table \ref{desiderata_table}, Row 6, while systems are expected to predict the true answer for known knowledge, as they contain the relevant context within their parameters, the ideal system would answer incorrectly/“unanswerable” for unknown knowledge given irrelevant context.



We follow the work by \citet{control, xie2023adaptive, roberts-etal-2020-much} 
and define known knowledge as questions that a model can answer correctly without context, and unknown knowledge as those it cannot.

\paragraph{Proposed Desiderata}
An ideal model should:

\begin{compactenum}[a.]
   \item[$\blacksquare$] For both known and unknown knowledge:
     \item \textit{Answer correctly with the original context}: this is the standard QA systems evaluation approach.\label{a}
     \item \textit{Answer correctly with a noisy irrelevant variation of the original context}: QA systems should be robust to distractors, as different users and information retrieval (IR) systems introduce varying amounts of irrelevant information.\label{b} 
     \item \label{c} \textit{Change its answer with conflicting context to the conflicting knowledge}: As our world is constantly changing,
     QA systems should be dynamic in their knowledge. That is, similarly to \citet{zhou2023contextfaithful, control}, we believe that the context should \textit{always} take priority over a model's parametric knowledge, when relevant. 
   \item[$\blacksquare$] For known knowledge:
    \item \textit{Answer correctly with no context}: In our setting this happens by default for known knowledge, as by definition known knowledge is questions that can be answered without context. However, we expect the ideal system to have the largest possible amount of knowledge, i.e., to be able to answer most questions without context.\label{d}
    \item \textit{Answer correctly with an irrelevant context}: Since the model answers questions correctly without context for known knowledge, it should also answer correctly with irrelevant context.\label{e}
   \item[$\blacksquare$] For unknown knowledge:
    \item \textit{Answer incorrectly/“unanswerable” with no context}: In our setting this happens by default for unknown knowledge.
    \textit{While the ideal model should predict “unanswerable” for questions it cannot answer}, 
    most existing datasets do not include questions that, according to our definition, are truly “unanswerable,” as they \textit{can} be answered with parametric knowledge (cf. Sec. \ref{Irrelevant_Knowledge}). Hence, we add here that models may also predict an incorrect answer, as expected from models that are forced to predict any answer other than “unanswerable” for unknown knowledge. \label{f}
    \item \textit{Answer the same with irrelevant context as with no context}:
    The ideal model should be consistent in its answer, even when wrong. Hence, the model's answer with irrelevant context and no context -- (f) above -- should be the same.\label{g} 
    \end{compactenum}

\section{Survey of Prior Work}
\label{related_work}

\subsection{Known vs. Unknown Knowledge}
As mentioned in Sec. \ref{ideal_QA},
the ideal behavior depends on if the knowledge contained in the context is known or unknown to the model. While most work evaluate on the entire data without such distinction, some analyze the known knowledge split:
\citet{xie2023adaptive, shaier-etal-2024-say, control} analyze models 
using a closed-book setting,
\citet{dist} assume the original contexts are known knowledge,
and \citet{chen-etal-2022-rich} evaluate correctly answered questions.

\subsection{The Standard Approach}
Row 1 in Table \ref{desiderata_table} shows the standard approach for evaluating QA systems, where systems are tasked with answering questions using a fixed context. For lack of space and since the focus of our survey is not the standard approach, we refer interesting readers to \citet{mcsurvey} and \citet{dzendzik-etal-2021-english} for further reading.

\subsection{Context + Distractor}
\label{Context_Distractor}
Next, we focus on Row 2 in Table \ref{desiderata_table}: the original context with a distractor, which measures the robustness of systems to various types of irrelevant (but not conflicting) noise. 

\paragraph{Overview}
Many analyze the susceptibility of QA systems to context-based noise. 
\citet{jia-liang-2017-adversarial} propose 
adding sentences that look similar to questions or random distractor words, which result in over $50\%$ decrease in performance. However, \citet{wang-bansal-2018-robust} mention that such unnatural distractors 
allow models to easily distinguish them and ignore them. 
Instead, they modify their approach by changing the locations of the distractors in addition to adding more fake answers.
\citet{si2019does} also modify the approach by further shuffling the distractor and find that BERT's performance drops by $50\%$. 
\citet{maharana-bansal-2020-adversarial} propose three new methods to generate QA adversaries which result in up to $45\%$ performance drop, while 
\citet{sen-saffari-2020-models} use context shuffling and find that the F1 scores of models decrease slightly. \citet{cao2022tasa} generate fluent and grammatical adversarial contexts which lower model confidence on the gold answer or direct the model towards an incorrect answer, and \citet{si-etal-2021-benchmarking} use character swapping and paraphrasing and show that state-of-the-art models are vulnerable. 
\citet{10066742} use random, structural, and irrelevant noise, and find that a sufficient amount of noise
can reduce the performance by $70\%$. \citet{liang2022holistic} focus on typos, such as capitalization or common misspellings, while \citet{schlegel2021semantics} use adverb modifications and find that models struggle with most of them.
Lastly, \citet{shi2023large} add an irrelevant sentence to the context which results in a dramatic decrease model performance.

The discussed work highlight that: 1) models can be easily dissuade by many types of distractors, even those that are nonsensical;
and 2) the type and complexity of the distractor matter and can result in either minimal or 
substantial performance drop.

\paragraph{Proposed Approaches}
\label{noise_Proposed_Approaches}
A popular approach to improve models' robustness to distractors is to train with augmented noisy data \cite{ribeiro-etal-2018-semantically, wang-bansal-2018-robust, maharana-bansal-2020-adversarial, bartolo-etal-2020-beat, michel-etal-2019-evaluation, gan-ng-2019-improving, moon-fan-2020-ask, cao2022tasa, si-etal-2021-benchmarking, khashabi-etal-2020-bang, control}. But some suggest that this has limited benefits \cite{jia-liang-2017-adversarial, wang-bansal-2018-robust,si-etal-2021-benchmarking}. 
Another possibility is to train models to edit distractor information, as done in 
\citet{bao-etal-2021-defending}, 
or to prompt systems to ignore irrelevant information \cite{shi2023large}.

\subsection{Conflicting Knowledge}
\label{Conflicting_Knowledge_survey}
Next, we focus on Rows 3 and 4 in Table \ref{desiderata_table}: contexts with information that is conflicting with the original context. The question is typically: how susceptible are systems to contexts that conflict with their parametric knowledge? 
While the alternate context conflicts with models' parametric knowledge in the known knowledge split, this is not necessarily the case for the unknown knowledge split, as the alternate context may already be contained within the model's parametric knowledge. 


\paragraph{Overview}
The most popular approach to evaluate systems on conflicting knowledge 
is entity substitution.
\citet{longpre-etal-2021-entity} replace the original answer entity with either a similar type one, an alias, an entity from the same corpus, or an entity based on popularity. This allows them to discover many aspects that affect models' over-reliance on their parametric knowledge, such as their size and domain. \citet{zhou2023contextfaithful} use a similar approach and focus on improving the robustness of systems to conflicting knowledge using prompts. \citet{chen-etal-2022-rich} modify the approach and use multiple contexts, and find that the performance of the IR system has a large effect on whether a model will use parametric knowledge.
\citet{dist} use the same approach 
but focus on disentangling systems' parametric and contextual knowledge,
while \citet{hong2023discern} find that models are very brittle to conflicting information in both in-context few-shot learning and fine-tuning settings.
\citet{eisenstein2022honest} find that models are approximately $3-4$ F1 points worse with conflicting entities, but also mention that such substitution can also affect the context's grammar.
\citet{yan-etal-2022-robustness} propose to use entities of different implications,
while, \citet{gardner-etal-2020-evaluating} find that models' performance can be reduced by up to $25\%$ with conflicting 
entities. 

The second most popular approach is to use negations.
\citet{gubelmann-handschuh-2022-context} automatically create contexts that are pragmatical specifically for each Transformer model, and find that most models are sensitive to negation. \citet{sen-saffari-2020-models} find that models continuously predict the original answer with 
negations,
and \citet{kassner-etal-2021-beliefbank} find that models often think that negative facts are true. Other methods also exist, such as using Mechanical Turkers \cite{contra} or graduate students \citet{varshney2023nlp}, which result in a significant performance change.
Some also use a masked language model to create conflicting knowledge \cite{contra, li-etal-2020-bert-attack}, where the former find that models are vulnerable to contradicting contexts, the latter mention that such an approach results in fluent and semantically preserving context. \citet{pan2023risk} use GPT-3.5 to generate conflicting contexts which result in a significant decline in system performance, while \citet{control} use T5 \cite{t5} and find that model's robustness does not scale with a model's size increase. \citet{zhong2023mquake}  randomly replace objects and find that models fail 
on conflicting multi-hop questions, while \citet{qian-etal-2022-perturbation} train a neural perturbation model to modify demographic terms. Lastly, \citet{gardner-etal-2020-evaluating} change the order the events or dates and find that model performance is greatly reduced.

The discussed work highlight that: 1) systems over-rely on their parametric knowledge, which often result in knowledge conflicts; 2) the type of conflicting information matters, but not necessarily for the right reasons. For example, 
\citet{eisenstein2022honest} find that entity substitution can affect the context's grammar, which can in general result in a decrease in performance.

\paragraph{Proposed Approaches}
As our knowledge is changing \cite{shaier2024adaptivequestionansweringenhancing}, \citet{mod} propose the task of modifying factual knowledge specifically in Transformer models \cite{transformer}, while \citet{edit} use a hyper-network to predict the weight update of systems.
\citet{fast_edit} use a collection of auxiliary networks that update a pretrained model’s behavior, and \citet{gpt_edit} identify factual-relevant neuron and update their weights. 
\citet{hong2023discern, pan2023risk, contra} propose a misinformation detector, but the latter mention that the benefits are limited with insufficient training data. \citet{xie2023adaptive} mention that improving the coherence of the context can improve the receptiveness of LMs to it, while \citet{longpre-etal-2021-entity} suggested to use a perfect retriever or to augment the training data with conflicting knowledge. \citet{khashabi-etal-2020-bang, qian-etal-2022-perturbation, control, varshney2023nlp, fang2023getting, chen-etal-2022-rich} also suggest to train with data augmentation, but the latter mention that
it does not easily generalize to other methods of creating conflicting knowledge. \citet{si2023prompting, zhou2023contextfaithful, pan2023risk} mention that carefully designed prompting strategies can improve the performance, while \citet{dist} suggest that models should generate two answers – a parametric one and a contextual one. \citet{zhong2023mquake} propose to store all edited facts externally, while \citet{yan-etal-2022-robustness} propose entity-based masking. Lastly, \citet{fortierdubois2023using} propose to use a natural language inference component to detect contradiction.

\subsection{Models' Parametric Knowledge}
\label{Parametric_Knowledge}
Next, we focus on Row 5 in Table \ref{desiderata_table}: an empty context with no distractor. This is the standard setting for evaluating model-internal knowledge or for determining whether models are “knowledge bases” \cite{petroni}. The question is: which facts are known or unknown to the model? 

\paragraph{Overview}
Recently, the size of LMs, which  are the basis of recent state-of-the-art QA models, has been increasing dramatically \cite{transformer, radford2018improving,radford_language_2019,palm,flan}. This in turn allows them to remember a massive amount of factual knowledge \cite{petroni, geva, roberts-etal-2020-much, kassner-schutze-2020-negated, editing, sung-etal-2021-language, shaier-etal-2024-comparing, shaier-etal-2023-emerging, shaier2023stochasticparrotsimitatingtell, jiang2020know, shin-etal-2020-autoprompt}. 
There are several ways to evaluate a model's parametric knowledge. For example, \citet{zhong-etal-2021-factual, shin-etal-2020-autoprompt, kassner-schutze-2020-negated, sung-etal-2021-language, petroni, jiang-etal-2020-know, dhingra-etal-2022-time, onoe-etal-2022-entity} use “fill in-the-blank” cloze statements, \citet{control, xie2023adaptive, roberts-etal-2020-much} use a closed-book setting,
and \citet{crawling} expand a knowledge graph around a seed entity by prompting the system.

The success of such models to recall factual information allows them to be useful in tasks that require knowledge, without supplying them with actual context \cite{how_much}, and even becoming competitive with other state-of-the-art fine-tuned models  \cite{few_shot}. However, training systems to memorize facts may also have adverse results. Systems have been shown to often ignore the context and focus on their parametric knowledge \cite{longpre-etal-2021-entity, kaushik-lipton-2018-much, ignore}. This results in hallucinations \cite{longpre-etal-2021-entity}, and poor performance 
when the knowledge is different than the training data
\cite{control, dist, longpre-etal-2021-entity}.

The discussed work highlight that: 1) there is no one correct approach to evaluate systems' knowledge;
2) developing systems with more 
knowledge is not necessarily better. For example, in domains where knowledge is often changing, it might be more important for systems to be more flexible to different contexts than knowledgeable, such as in medicine, where new treatments often arise.
\paragraph{Proposed Approaches}
While many evaluate parametric knowledge, not many \textit{directly} focus on increasing it. However, existing experiments show that bigger models or different architectures can help \cite{petroni, roberts-etal-2020-much}. Furthermore, better knowledge can also be learned via multimodal training \cite{arocaouellette2021prost}.

\subsection{Irrelevant Knowledge}
\label{Irrelevant_Knowledge_overview}
Next, we focus on Row 6 in Table \ref{desiderata_table}: irrelevant context. The question is: how often does a system changes its answers when given irrelevant context? 

\paragraph{Overview}
What we define as irrelevant context exists in many datasets, such as the Natural Questions \cite{kwiatkowski-etal-2019-natural}, SQuAD 2.0 \cite{rajpurkar-etal-2018-know}, QuAC \cite{choi-etal-2018-quac}, CoQA \cite{reddy-etal-2019-coqa}, and MS MARCO \cite{msmacro}, where the answer to the question is not supported by the context. 
Other work also evaluate such irrelevant context formulation, such as \citet{control}, which define irrelevant contexts as those that do not entail the answer. They find that models are strongly interfered by irrelevant contexts, especially those that share a similar general topic as the question. Additionally, \citet{dist} find that random irrelevant context is more challenging to models in some settings. 

As the field is moving towards using LLMs which contain large amount of knowledge, these type of questions become not truly “unanswerable” with irrelevant context, or even without any context, which further reinforce the need to split existing evaluations into known and unknown knowledge. This is in comparison to subjective, philosophical, or imagination questions such as proposed in \citet{know?}, which are truly “unanswerable” by any system, regardless of their knowledge.

\paragraph{Proposed Approaches}
While we discuss many approaches to improve model robustness on contexts with added distrators in Section \ref{noise_Proposed_Approaches}, not many evaluate models using irrelevant context of our setting -- based on known and unknown knowledge. \citet{control, dist} propose training with data augmentation, while the latter 
further train the model to disentangle its parametric and contextual knowledge by generating two answers.

\section{Defining Desiderata}
\subsection{Problem Formulation}
\label{Problem_Formulation}
Given a dataset composed of questions ${q_1, q_2, ... , q_n}\in Q$, their corresponding answers ${a_1, a_2, ... , a_n}\in A$ and contexts ${c_1, c_2, ... , c_n}\in C$, 
we evaluate how well a model uses the context or its modifications, which will be presented in the following sections, on the given questions.

We note that two types of context-based QA datasets exist: 1) the questions are about a general knowledge concept and the contexts supplement the knowledge, for example, \textit{"Who is the current president?"} Relevant datasets are, e.g., WikiQA \cite{yang-etal-2015-wikiqa}, SQuAD 1.0 \cite{squad}, and OpenBookQA \cite{openbookqa}; and 2) the questions are specifically about the contexts, for example, \textit{"What did the narrator mean by [...]"}. Datasets include, e.g., Race \cite{race}, QuAIL \cite{quail}, and CosmosQA \cite{cosmos}. \textbf{We only use the first type}, as those questions can be used to measure models' parametric knowledge by omitting the context, while the latter cannot, as without context the models cannot answer the question.\footnote{If they do, it is by mere chance.}

\subsubsection{Creating Conflicting Context} 
\label{conflicting}
We 
follow \citet{contra, li-etal-2020-bert-attack}'s approach of using a masked language model. More formally, we mask the answer string $a_i$ from the context $q_i$ when it exists verbatim.\footnote{The answer string is sometimes paraphrased in the context. We discard such questions.} We then use DistilBERT \cite{sanh2020distilbert} to predict the masked answer, and replace the masked token with it. For each masked answer we generate 10 different answers, and remove any that are similar (i.e., exact string match) to the original answer. 
This results in up to 10 conflicting contexts for each question. In the free-form setting, we then replace the original answer $a_i$ with the new predicted answer. For the MC setting, we leave the original answer as one of the MC options and replace one wrong answer with the new answer. The ideal behavior of systems for such context can be seen in Table \ref{desiderata_table} in Rows 3 and 4.


\subsubsection{Creating Irrelevant Context} 
\label{Irrelevant_Knowledge}


What we define as irrelevant context exists in many datasets, such as those described in Section \ref{Irrelevant_Knowledge_overview}.
These type of questions have been termed “unanswerable” questions. However, in our formulation, if the context is irrelevant or does not exist, models may still have the parametric knowledge to answer the corresponding question (e.g., from pretraining), which makes these questions not truly “unanswerable.” \textbf{We avoid using such datasets as the correct answer is not provided} (e.g., SQuAD 2.0 has empty strings as labels for its irrelevant contexts), which prevents us from determining if the model has the parametric knowledge to answer. 

To create irrelevant context we opt for a method that can be applied to most existing context-based QA dataset and follow \citet{dist}'s approach of selecting random contexts. More formally, for each question $q_i$, we replace the corresponding context $c_i$ with a random context $c_j \in C$, where $c_i \neq c_j$. We repeat this 5 times which results in 5 irrelevant contexts for each question.\footnote{While there is a small chance that the random context contain some relevant information, it is unlikely.} The ideal behavior of systems for such context can be seen in Table \ref{desiderata_table} in Row 6.


\subsubsection{Context with Distractor}
\label{noisy}
To add a distractor to contexts
we use the ADDANY approach \cite{jia-liang-2017-adversarial}, 
but modify it to be applicable to free-form and MC settings. 
In particular, instead of modifying $w_i$ to be the $x$ that minimizes the expected value of the F1 score, we 
update it
to be the one that maximizes the perplexity of the answer with respect to the input string in the free-form setting, and the one which minimizes the probability of the correct answer for MC. 


\section{Experiments}
In our experiments, each context $c_i$ and question $q_i$ are input into the model within the following string: “question: $q_i$. context: $c_i$.” In the free-form setting and for MCQA, we use exact match (EM) and, respectively, accuracy to measure model performance. That being said, our approach is by design extremely easily adaptable to different choices of metrics, such as LLM-based ones \cite{kamalloo-etal-2023-evaluating}, which could have a higher correlation with humans for QA tasks than EM.

\subsection{Datasets}
We experiment on 5 datasets: 1) SQuAD 1.0 \cite{squad},\footnote{We have two annotators perform a manual analysis of a subset of 100 SQuAD questions to determine the percentage of questions that are uniquely answerable without context (see Appendix \ref{squad_manual}).}$^,$\footnote{The reason we use SQuAD 1.0 and not the later version is discussed in Section \ref{Irrelevant_Knowledge}.} 2) AdversarialQA \cite{bartolo-etal-2020-beat}, which we both use for free-form QA, where the answer span is to be generated, as well as 3) Natural Questions \cite{kwiatkowski-etal-2019-natural}. Additionally, we also use 4) SciQ \cite{Welbl2017CrowdsourcingMC} and 5) MedMCQA \cite{pal2022medmcqa}, which are MCQA datasets. For further datasets statistics see Appendix \ref{dataset_stats}.

\subsection{Models}
\label{models}
We evaluate 5 LLM-based QA models in the free-form setting: GPT 3.5 \cite{openai_2023}, GPT 4 \cite{openai2023gpt4}, BART \cite{BART} base,
T5 \cite{t5} small,
and six LLM-based QA models in the MCQA setting: BERT \cite{bert} base,
BigBird \cite{bigbird} base,
Longformer \cite{longformer} base,
RoBERTa \cite{liu2019roberta} base,
ALBERT \cite{lan2020albert} base, 
and DistilBERT \cite{sanh2020distilbert} base.
We finetune each pretrained model on the training set for 20 epochs, use early stopping on the validation with patience of 3, and evaluate them on the test set. As the test sets for SQuAD 1.0 and Natural Questions are not publically available, we split the validation set into 2 for all models, and use one half as the test set. 
Lastly, on the Natural Questions dataset we evaluate 3 published models (without further training) from \citet{roberts-etal-2020-much} to analyze how they score on our desiderata. These models are 1) T5-Small,\footnote{https://huggingface.co/google/t5-small-ssm-nq} 2) T5-Large-1.0,\footnote{T5 large that is fine-tuned on $100\%$ of the train splits of Natural Questions. https://huggingface.co/google/t5-large-ssm-nq} and 3) T5-Large-0.9.\footnote{T5 large that is fine-tuned on $90\%$ of the train splits of Natural Questions. https://huggingface.co/google/t5-large-ssm-nqo} The results can be seen in Appendix \ref{NaturalQuestions_results}.

\subsection{Results and Analysis}
Our toolkit takes most context-based datasets, as described in Section \ref{Problem_Formulation}, and automatically prepares and evaluates all desiderata aspects \textit{at once}. We use it to evaluate each of the models described in Section \ref{models} in all of the settings shown in Table \ref{desiderata_table}. In comparison to previous work, we split desiderata aspects by finding the context that is known and unknown to individual models, as the ideal behavior of models' depends on if the knowledge contained in the context is known or unknown to the model.
Our results can be seen in Tables \ref{results_table_MC} and \ref{results_table_FF}.

\begin{table*}[t]
\centering
\setlength{\tabcolsep}{3.0pt}
\include{new_table2}
\caption{Results table: MCQA models. K. Am=Knowledge amount; St=Standard; KK=known knowledge; UK=unknown knowledge; Dist=distractor; Conf=conflicting; Irr=Irrelevant. Each model's parametric knowledge results in different known and unknown knowledge splits which we evaluate using accuracy. In bold, highest accuracy on each of the desiderata components for each dataset.}
\label{results_table_MC}
\end{table*}
\begin{table*}[t]
\centering
\setlength{\tabcolsep}{3.0pt}
\include{new_table3}
\caption{Results table: free-form models. K. Am=Knowledge amount; St=Standard; KK=known knowledge; UK=unknown knowledge; Dist=distractor; Conf=conflicting; Irr=Irrelevant. Each model's parametric knowledge results in different known and unknown knowledge splits which we evaluate using accuracy. In bold, highest accuracy on each of the desiderata components for each dataset. The distractor setting is not done for the GPT models as it requires model accesss.}
\label{results_table_FF}
\end{table*}

\paragraph{Amount of Knowledge}
We calculate the amount of knowledge models possess using the closed-book setting and accuracy, as described in Section \ref{Problem_Formulation}. On the SciQ and MedMCQA datasets, models possess sufficient knowledge to accurately respond to approximately half and one-third of all queries, respectively, without additional context. Interestingly, ALBERT performs the poorest on both datasets, achieving an accuracy rate of $45.3\%$ on SciQ and $22.7\%$ on MedMCQA. In contrast, BigBird and Longformer score the highest on SciQ and MedMCQA, with accuracies of $56.4\%$ and $32.3\%$, respectively. This aligns with previous discussed work in Section \ref{Parametric_Knowledge}, which suggest that such models contain abundant factual information and have the potential to be used as open-domain QA systems. 

In comparison, the free-form models could not answer even $9\%$ of the questions successfully without context (GPT-4 scores $8.7\%$ on SQuAD).\footnote{Due to the small number of correct instances, we cannot draw any strong conclusions regarding such systems in the known vs. unknown knowledge splits.}$^,$\footnote{We also have two annotators perform a manual analysis of a subset of GPT 3.5 outputs, see Appendix \ref{chatgpt_manual}.} The significant difference in performance between the MC and the free-form models may partially be due to the fact that the MC setting is much easier, where a model that randomly predicts an answer gets on average $25\%$ of the questions correctly.\footnote{We also try non-finetuned versions of the free-form models, but the results are comparable.} 

\paragraph{The Standard Evaluation} 
Almost all models (except for ALBERT on MedMCQA and T5-small on SQuAD) score higher on the known vs. the unknown knowledge split. For example, $99.1\%$ vs $96.6\%$ for BigBird on SciQ and $58.4\%$ vs $4.8\%$ for GPT-4 on AdversarialQA. This suggests that models find context that reinforce their knowledge beneficial, which 
emphasize that future work should evaluate systems from knowledge perspective. 

\paragraph{Distractor}
Similar to previous work (cf. Sec \ref{Context_Distractor}), we find a significant reduction in performance across all MC models (e.g., on SciQ, DistilBERT's performance drops from $97.4\%$ to $4.0\%$ on known knowledge). Furthermore, the difference between known and unknown knowledge is visible, where across almost all models (except for DistilBERT on SciQ, and Longformer and ALBERT on MedMCQA) noise affect unknown knowledge more. 
While there is also a clear reduction in performance for free-form models, the reduction is not as large. For example, T5 small drops from $72.6\%$ in the unknown knowledge split to $68.9\%$. 

\paragraph{Conflicting Knowledge}
We also find a substantial performance drop across all models when conflicting knowledge is introduced. For example, $33.2\%$ for RoBERTa in the known knowledge split on SciQ, and $50.0\%$ for GPT 3.5 in the known knowledge split on AdversarialQA. We also find again, a difference in behavior across almost all MC models between known and unknown knowledge: the performance drop is lower in the unknown split, which we believe occurs as, for the known knowledge split, this type of substitution conflicts with the model's parametric knowledge, while this might not be the case for the unknown split as discussed in Sec \ref{Conflicting_Knowledge_survey}. 

\paragraph{Irrelevant Context}
We find that all models are more consistent with their answers for known knowledge when irrelevant contexts are added. For example, T5-base generates similar answers to $65.1\%$ of the questions for known knowledge and only $21.3\%$ to questions for unknown on SQuAD, while Longformer generates similar answers to $53.5\%$ vs $50.2\%$ for known and unknown knowledge on MedMC, respectively. This might suggest that systems are more confident about known information and hence are less likely to change answers. 

\paragraph{Distractor + Conflicting Knowledge Combined}
Looking at the \textit{combination} of distractors with conflicting contexts, we find that the performance drop is generally lower in the unknown split for most models. 
We can also see that the combination of conflicting contexts and added distractor can result in accuracy drop of close to $96\%$, such as in DistilBERT 
in known knowledge 
on SciQ. 

\paragraph{Distractor + Conflicting Knowledge -- Separate}
Looking at the models' performances in the conflicting knowledge and distractor addition settings \textit{separately}, we can further see that systems that are more susceptible to noise are often more likely to correctly answer according to a context that conflicts with their parametric knowledge. For example, within the MC systems, DistilBERT has the largest performance decrease for added distracor, but also performs nearly the best on conflicting knowledge on SciQ. Similar trends can be seen between ALBERT and RoBERTa, Longformer and RoBERTa, BERT and BigBird, and others. A potential reason might be that the susceptibility of systems to noise occurs as they are more attentive to everything in the context, which is beneficial for conflicting knowledge.

\paragraph{Distractor + Consistency}
Looking at models' performances for the distractor and irrelevant context settings, we find that systems that are less susceptible to distractors are not necessarily more consistent with their answers when provided irrelevant context. BigBird is the more susceptible to distractors than Longformer on SciQ, and less consistent than it for unknown data, where opposite trends occur between BigBird and Longformer. 

\paragraph{MC vs. Free-form}
For added distractors, we find that MC models are more susceptible than the free-form ones, and have a larger performance drop. This may be due to the fact that such models are less susceptible to noise, or that the optimization method we use to find noisier sentences in the free-form is not as strong as the one we apply in the MC setting (Section \ref{noisy}). For conflicting knowledge, the reduction in performance between the MC models and the free-form ones is also visible and somewhat comparable. For example, GPT-4 score is reduced by $53.9\%$
on the known knowledge split when conflicting knowledge is added on AdvarsarialQA, in comparison to BigBird's performance on MedMCQA decreases by $36.0\%$.

\paragraph{Model Size}
We also test similar types of models in two sizes: T5-small and T5-base, and GPT-3.5 and GPT-4. We find that the larger variant 1) has a larger amount of known knowledge. For example, the T5 models score $0.9\%$ vs $0.3\%$ on SQuAD and $4.2\%$ vs $2.9\%$ on AdvarsarialQA, where the GPT models score $8.7\%$ vs $0.3\%$ on SQuAD and $5.9\%$ vs $0.2\%$ on AdvarsarialQA; 2) is more robust to distractors. For example, T5-base decreases by $6.0\%$ 
on known knowledge on SQuAD, where the smaller version decreases by $10.0\%$;
3) is not necessarily more robust to conflicting knowledge on known knowledge. For example, GPT-4's performance drop is larger than GPT-3.5 on SQuAD, but T5-base's drop is lower than T5-small on the same dataset; 4) is not necessarily more consistent with its answers. For example, T5-small is more consistent for unknown knowledge on SQuAD and AdvarsarialQA, but less consistent for known knowledge. Oppositely, GPT-4 is more consistent for unknown knowledge on AdvarsarialQA, but less consistent on SQuAD.

\section{Conclusion}
We outline a set of -- previously discussed as well as novel -- desiderata for context-based QA systems. We survey relevant papers to provide an overview of the state of the field, and evaluate 15 QA systems according to all desiderata \textit{at once}.  While previous work examine desiderata aspects in isolation, by looking at all aspects together, we are able to find novel trends which increase our understanding of how these models work and reveal potential avenues to improve such models.

\section*{Limitations}
While we try to be comprehensive in the survey and cover many existing influential work, we may have missed some for the large number of them. However, we believe that this should not influence the found trends. Additionally, as a major part of our analysis is based on splitting the data into the known and unknown based on models' parametric knowledge, it is important to note that currently no perfect approach to measure parametric knowledge exist. Hence, the results may be slightly skewed, as for example, models may have guessed on questions in our closed-book formulation which resulted in more questions in the known data split. 



\section*{Ethics Statement}
The motivation for this paper is to unify many existing aspects from QA systems so that we can find trends and have a better evaluation strategy of such models. We believe that it is crucial that future work continues to evaluate and improve model robustness so they can be safely used in practical scenarios. 

\section*{Acknowledgments}
We thank the reviewers for their comments and great suggestions. The authors acknowledge financial support from NIH grants OT2TR003422 and R01LM013400.



\bibliography{anthology,custom}
\bibliographystyle{acl_natbib}

\clearpage 

\appendix
\section{Datasets Statistics}
\label{dataset_stats}
\paragraph{SQuAD 1.0}
The Stanford Question Answering Dataset (SQuAD) \cite{squad} is a widely used benchmark for evaluating reading comprehension systems. The version 1.0 release contains over 100k crowd-sourced question--answer pairs with more than 500 Wikipedia articles. The questions were created by humans who were instructed to submit up to five questions on the contexts of the passage they had read. The answer to each question is a segment of text from the corresponding reading passage. In our setting, we reformulate the original task of predicting a text segment on the context into free-form text generation.

\paragraph{AdversarialQA}
The AdversarialQA dataset \cite{bartolo-etal-2020-beat} contains 36k questions which were created using three different models in the annotation process. The annotation approach entails humans formulating questions designed to test current QA models, deliberately crafting queries that these models struggle to answer accurately. These questions are then used for annotating the dataset, resulting in samples collected through adversarial means.

\paragraph{Natural Questions}
The Natural Questions dataset \cite{kwiatkowski-etal-2019-natural} contains about 320k questions, which were created using real users' queries from the Google search engine. Every question is linked to a Wikipedia page from the top-5 search outcomes, and annotators produce a long response and a concise response if they are available on the page.

\paragraph{MedMCQA}
The Medical Multiple-Choice Question Answering (MedMCQA) dataset \cite{pal2022medmcqa} contains 194k MCQA questions in the medical domain, around 21 medical subjects. The questions require deep language understanding, as they assess 
models' reasoning capabilities across various medical subjects and topics, encompassing over ten different types of reasoning skills.

\paragraph{SciQ}
The SciQ dataset \cite{Welbl2017CrowdsourcingMC} contains about 13k science exam questions about chemistry, physics, biology, and more. The questions are in MC format, each with four answers where only one is correct. Most of the questions contain an additional context paragraph with supporting evidence for the correct answer. In our setting, we discard questions that do not. 

\begin{table*}[th]
\centering
\small
\setlength{\tabcolsep}{3.0pt}
\begin{tabular}{|c|c|c|c|c|c|c|c|c|c|c|c|c|c|}
\hline
\textbf{Dataset}                           & \textbf{Model} & \textbf{\begin{tabular}[c]{@{}c@{}}K.\\ Am.\end{tabular}} & \textbf{\begin{tabular}[c]{@{}c@{}}St.\\ KK\end{tabular}} & \textbf{\begin{tabular}[c]{@{}c@{}}St. \\ UK\end{tabular}} & \textbf{\begin{tabular}[c]{@{}c@{}}St. \\ Avg\end{tabular}} & \textbf{\begin{tabular}[c]{@{}c@{}}Dist.\\ KK\end{tabular}} & \textbf{\begin{tabular}[c]{@{}c@{}}Dist.\\ UK\end{tabular}} & \textbf{\begin{tabular}[c]{@{}c@{}}Conf.\\ KK\end{tabular}} & \textbf{\begin{tabular}[c]{@{}c@{}}Conf.\\ UK\end{tabular}} & \textbf{\begin{tabular}[c]{@{}c@{}}Conf.\\ Dist.\\ KK\end{tabular}} & \textbf{\begin{tabular}[c]{@{}c@{}}Conf.\\ Dist.\\ UK\end{tabular}} & \textbf{\begin{tabular}[c]{@{}c@{}}Irr.\\ KK\end{tabular}} & \textbf{\begin{tabular}[c]{@{}c@{}}Irr.\\ UK\end{tabular}} \\ \hline
\multirow{3}{*}{\textbf{NaturalQuestions}} & T5-Small       & 11.9                                                      & 47.4                                                      & 6.0                                                        & 10.9                                                        & 37.5                                                        & 4.1                                                         & 2.5                                                         & 3.2                                                         & 0.0                                                                 & 1.0                                                                 & 10.0                                                       & 2.2                                                        \\ \cline{2-14} 
                                           & T5-Large-0.9   & 16.9                                                      & 84.5                                                      & 21.3                                                       & 32.0                                                        & 67.3                                                        & 28.1                                                        & 10.7                                                        & 9.9                                                         & 10.3                                                                & 6.9                                                                 & 51.9                                                       & 23.7                                                       \\ \cline{2-14} 
                                           & T5-Large-1.0   & 18.2                                                      & 85.9                                                      & 21.0                                                       & 32.9                                                        & 80.0                                                        & 27.2                                                        & 10.5                                                        & 9.8                                                         & 7.1                                                                 & 6.5                                                                 & 44.6                                                       & 22.3                                                       \\ \hline
\end{tabular}
\caption{Results table: free-form models on the Natural Questions dataset. K. Am=Knowledge amount; St=Standard; KK=known knowledge; UK=unknown knowledge; Dist=distractor; Conf=conflicting; Irr=Irrelevant. Each model's parametric knowledge results in different known and unknown knowledge splits which we evaluate using accuracy.}
\label{NQ_table}
\end{table*}
\section{Manual Analysis of SQuAD}
\label{squad_manual}
As noted in Section \ref{Problem_Formulation}, two types of context-based QA datasets exist, and we only use the first type as those questions can be used to measure models’ parametric knowledge by omitting the context. To that end, two annotators perform a manual analysis of a subset of 100 SQuAD questions to see what percentage of questions are uniquely answerable without context and find that $69\%$ of the questions can be answered without the context.

\section{Natural Questions Results}
\label{NaturalQuestions_results}
We additionally evaluate 3 published models from \citet{roberts-etal-2020-much} on the Natural Questions dataset to analyze how they score on our desiderata. The results can be seen in Table \ref{NQ_table}. However, in comparison to \citet{roberts-etal-2020-much}, which omitted the questions corresponding to the “unanswerable” labels and long answers, as they “are nearly impossible to predict without the oracle context,” we evaluate on the entire set.

\section{Manual Analysis of ChatGPT}
\label{chatgpt_manual}
While we use exact match in our experiments, two annotators manually evaluate 100 generated responses from GPT 3.5 to analyze how many of the generated responses actually answer the questions (i.e., not using exact match) and find that number to be $28\%$. This is significantly higher than the exact match scores, which highlights that exact match may not be the best method to analyze model responses. That being said, our approach is by design extremely easily adaptable to different choices of metrics, such as LLM-based ones \cite{kamalloo-etal-2023-evaluating}, which could have a higher correlation to humans than exact match for QA tasks.

\end{document}

%% file: new_table1.tex
\begin{table}[]
\centering\small\setlength{\tabcolsep}{2.5pt}
\begin{tabular}{l | c|c|c|c}
\toprule
& Context      & Distractor & \shortstack{Known\\Knowledge} & \shortstack{Unknown\\Knowledge} \\ \midrule
1& Original  
& & T                              & T                                 \\ 
2 & Original 
&  \checkmark & T                              & T                                 \\ 
3 & Alternative 
& & A                & A                                 \\ 
4 & Alternative
& \checkmark & A                & A                                 \\ 
5 & None         & & T  
& B 
\\ 
6 & Irrelevant & & T 

& B
\\ 
\bottomrule
\end{tabular}
\caption{Desiderata table: what should an optimal model do for different types of contexts? T = \textit{true answer}; A = \textit{conflicting answer}; B = \textit{wrong answer/unanswerable}; The distractor (cf. Sec. \ref{Context_Distractor}) is a string of words that is concatenated to the context (cf. Sec. \ref{noisy}); Alternative context (cf. Sec. \ref{Conflicting_Knowledge_survey}) is a slight modification of the original context, where we replace the answer string with an alternative one (cf. Sec. \ref{conflicting}); Irrelevant context (cf. Sec. \ref{Irrelevant_Knowledge_overview}) is a random context (cf. Sec. \ref{Irrelevant_Knowledge}).}
\label{desiderata_table}
\end{table}

%% file: new_table2.tex
\begin{tabular}{|c|c|c|c|c|c|c|c|c|c|c|c|c|c|}
\hline
\textbf{Dataset}                  & \textbf{Model} & \textbf{\begin{tabular}[c]{@{}c@{}}K. \\ Am. \end{tabular}} & \textbf{\begin{tabular}[c]{@{}c@{}}St.\\ KK\end{tabular}} & \textbf{\begin{tabular}[c]{@{}c@{}}St. \\ UK\end{tabular}} & \textbf{\begin{tabular}[c]{@{}c@{}}St. \\ Avg\end{tabular}} & \textbf{\begin{tabular}[c]{@{}c@{}}Dist.\\ KK\end{tabular}} & \textbf{\begin{tabular}[c]{@{}c@{}}Dist.\\ UK\end{tabular}} & \textbf{\begin{tabular}[c]{@{}c@{}}Conf.\\ KK\end{tabular}} & \textbf{\begin{tabular}[c]{@{}c@{}}Conf.\\ UK\end{tabular}} & \textbf{\begin{tabular}[c]{@{}c@{}}Conf.\\ Dist.\\ KK\end{tabular}} & \textbf{\begin{tabular}[c]{@{}c@{}}Conf.\\ Dist.\\ UK\end{tabular}} & \textbf{\begin{tabular}[c]{@{}c@{}}Irr.\\ KK\end{tabular}} & \textbf{\begin{tabular}[c]{@{}c@{}}Irr.\\ UK\end{tabular}} \\ \hline
\multirow{6}{*}{\textbf{SciQ}}    & BERT           & 52.7                                                                 & 97.4                                                           & 95.2                                                            & 96.3                                                             & 56.5                                                        & 46.3                                                        & 63.1                                                        & 69.7                                                        & 29.8                                                                & \textbf{34.9}                                                       & \textbf{82.8}                                              & \textbf{73.8}                                              \\ \cline{2-14} 
                                  & BigBird        & \textbf{56.4}                                                        & 99.1                                                           & 96.6                                                            & 97.9                                                             & 36.7                                                        & 23.6                                                        & \textbf{75.2}                                               & \textbf{78.6}                                               & 11.0                                                                & 13.1                                                                & 78.9                                                       & 60.4                                                       \\ \cline{2-14} 
                                  & Longformer     & 55.4                                                                 & \textbf{99.5}                                                  & \textbf{98.4}                                                   & \textbf{99.0}                                                    & \textbf{71.4}                                               & \textbf{61.5}                                               & 66.3                                                        & 71.4                                                        & \textbf{31.2}                                                       & 34.2                                                                & 81.4                                                       & 68.4                                                       \\ \cline{2-14} 
                                  & RoBERTa        & 51.9                                                                 & \textbf{99.5}                                                  & 96.7                                                            & 98.1                                                             & 20.0                                                        & 9.0                                                         & 73.4                                                        & 77.9                                                        & 17.0                                                                & 7.3                                                                 & 76.6                                                       & 65.1                                                       \\ \cline{2-14} 
                                  & ALBERT         & 45.3                                                                 & 99.2                                                           & 97.1                                                            & 98.1                                                             & 55.0                                                        & 43.7                                                        & 69.4                                                        & 74.6                                                        & 20.0                                                                & 25.9                                                                & 80.5                                                       & 71.4                                                       \\ \cline{2-14} 
                                  & DistilBERT     & 49.6                                                                 & 97.4                                                           & 94.8                                                            & 96.1                                                             & 4.0                                                         & 4.0                                                         & 73.0                                                        & 76.5                                                        & 1.0                                                                 & 1.0                                                                 & 67.9                                                       & 61.7                                                       \\ \hline
\multirow{6}{*}{\textbf{MedMC}} & BERT           & 31.1                                                                 & 84.1                                                           & \textbf{81.6}                                                   & 82.9                                                             & 75.7                                                        & \textbf{64.3}                                               & 56.5                                                        & \textbf{61.8}                                               & \textbf{73.5}                                                       & 61.3                                                                & 66.7                                                       & 61.6                                                       \\ \cline{2-14} 
                                  & BigBird        & 27.3                                                                 & 83.9                                                           & 74.5                                                            & 79.2                                                             & 65.5                                                        & 51.9                                                        & 47.9                                                        & 55.4                                                        & 21.3                                                                & 32.5                                                                & 58.8                                                       & 53.0                                                       \\ \cline{2-14} 
                                  & Longformer     & \textbf{32.3}                                                        & 84.9                                                           & 78.4                                                            & 81.7                                                             & 61.7                                                        & 61.1                                                        & 53.2                                                        & 55.0                                                        & 58.7                                                                & \textbf{70.6}                                                       & 53.5                                                       & 50.2                                                       \\ \cline{2-14} 
                                  & RoBERTa        & 28.7                                                                 & \textbf{88.3}                                                  & 81.2                                                            & \textbf{84.7}                                                    & \textbf{76.6}                                               & 60.5                                                        & 62.1                                                        & 60.6                                                        & 73.0                                                                & 64.6                                                                & 59.4                                                       & 51.7                                                       \\ \cline{2-14} 
                                  & ALBERT         & 22.7                                                                 & 76.6                                                           & 77.9                                                            & 77.3                                                             & 41.6                                                        & 62.1                                                        & 39.8                                                        & 42.3                                                        & 38.4                                                                & 58.1                                                                & 38.8                                                       & 34.8                                                       \\ \cline{2-14} 
                                  & DistilBERT     & 28.8                                                                 & 84.1                                                           & 76.6                                                            & 80.3                                                             & 63.3                                                        & 53.9                                                        & \textbf{62.5}                                               & 60.0                                                        & 40.5                                                                & 46.3                                                                & \textbf{66.9}                                              & \textbf{62.3}                                              \\ \hline
\end{tabular}

%% file: new_table3.tex
\begin{tabular}{|c|c|c|c|c|c|c|c|c|c|c|c|c|c|}
\hline
\textbf{Dataset}                  & \textbf{Model} & \textbf{\begin{tabular}[c]{@{}c@{}}K.\\ Am.\end{tabular}} & \textbf{\begin{tabular}[c]{@{}c@{}}St.\\ KK\end{tabular}} & \textbf{\begin{tabular}[c]{@{}c@{}}St. \\ UK\end{tabular}} & \textbf{\begin{tabular}[c]{@{}c@{}}St. \\ Avg\end{tabular}} & \textbf{\begin{tabular}[c]{@{}c@{}}Dist.\\ KK\end{tabular}} & \textbf{\begin{tabular}[c]{@{}c@{}}Dist.\\ UK\end{tabular}} & \textbf{\begin{tabular}[c]{@{}c@{}}Conf.\\ KK\end{tabular}} & \textbf{\begin{tabular}[c]{@{}c@{}}Conf.\\ UK\end{tabular}} & \textbf{\begin{tabular}[c]{@{}c@{}}Conf.\\ Dist.\\ KK\end{tabular}} & \textbf{\begin{tabular}[c]{@{}c@{}}Conf.\\ Dist.\\ UK\end{tabular}} & \textbf{\begin{tabular}[c]{@{}c@{}}Irr.\\ KK\end{tabular}} & \textbf{\begin{tabular}[c]{@{}c@{}}Irr.\\ UK\end{tabular}} \\ \hline
\multirow{5}{*}{\textbf{SQuAD}}   & T5-Small       & 0.3                                                       & 70.0                                                      & 72.6                                                       & 72.6                                                        & 60.0                                                        & 68.9                                                        & 53.1                                                        & 63.5                                                        & 53.1                                                                & 55.4                                                                & 45.9                                                       & \textbf{25.8}                                              \\ \cline{2-14} 
                                  & T5-Base        & 0.9                                                       & \textbf{82.0}                                             & \textbf{78.4}                                              & \textbf{78.4}                                               & \textbf{76.0}                                               & \textbf{75.4}                                               & \textbf{75.6}                                               & \textbf{64.7}                                               & \textbf{70.7}                                                       & \textbf{61.3}                                                       & \textbf{65.1}                                              & 21.3                                                       \\ \cline{2-14} 
                                  & BART           & 0.9                                                       & 68.7                                                      & 65.4                                                       & 65.4                                                        & 60.4                                                        & 59.9                                                        & 55.0                                                        & 50.2                                                        & 51.3                                                                & 43.0                                                                & 48.3                                                       & 24.0                                                       \\ \cline{2-14} 
                                  & GPT-3.5        & 0.3                                                       & 50.0                                                      & 0.3                                                        & 0.4                                                         & -                                                           & -                                                           & 33.3                                                        & 0.1                                                         & -                                                                   & -                                                                   & 2.7                                                        & 2.6                                                        \\ \cline{2-14} 
                                  & GPT-4          & \textbf{8.7}                                              & 45.3                                                      & 10.4                                                       & 13.4                                                        & -                                                           & -                                                           & 12.1                                                        & 6.5                                                         & -                                                                   & -                                                                   & 32.3                                                       & 0.6                                                        \\ \hline
\multirow{5}{*}{\textbf{Adv. QA}} & T5-Small       & 2.9                                                       & 63.6                                                      & 20.1                                                       & 21.4                                                        & 59.0                                                        & 19.3                                                        & 6.0                                                         & 16.5                                                        & 4.3                                                                 & 14.6                                                                & \textbf{69.6}                                              & \textbf{24.9}                                              \\ \cline{2-14} 
                                  & T5-Base        & 4.2                                                       & 65.0                                                      & \textbf{27.2}                                              & \textbf{28.8}                                               & 60.3                                                        & \textbf{37.7}                                               & \textbf{11.8}                                               & \textbf{19.2}                                               & \textbf{10.5}                                                       & \textbf{20.5}                                                       & 57.1                                                       & 5.4                                                        \\ \cline{2-14} 
                                  & BART           & 4.1                                                       & \textbf{87.0}                                             & 20.2                                                       & 23.0                                                        & \textbf{77.4}                                               & 16.7                                                        & 9.2                                                         & 11.8                                                        & 6.7                                                                 & 7.6                                                                 & 60.2                                                       & 13.6                                                       \\ \cline{2-14} 
                                  & GPT-3.5        & 0.2                                                       & 50.0                                                      & 2.4                                                        & \textbf{2.5}                                                & \textbf{-}                                                  & -                                                           & 0.0                                                         & 0.5                                                         & -                                                                   & -                                                                   & 50.0                                                       & 0.9                                                        \\ \cline{2-14} 
                                  & GPT-4          & \textbf{5.9}                                              & 58.4                                                      & 4.8                                                        & 8.0                                                         & -                                                           & -                                                           & 4.5                                                         & 1.5                                                         & -                                                                   & -                                                                   & 41.5                                                       & 11.9                                                       \\ \hline
\end{tabular}

%% file: main.bbl
\begin{thebibliography}{98}
\expandafter\ifx\csname natexlab\endcsname\relax\def\natexlab#1{#1}\fi

\bibitem[{Alexandrov et~al.(2023)Alexandrov, Zakharova, and Butakov}]{10066742}
Dmitriy Alexandrov, Anastasiia Zakharova, and Nikolay Butakov. 2023.
\newblock \href {https://doi.org/10.1109/ICSC56153.2023.00012} {Does noise really matter? investigation into the influence of noisy labels on bert-based question answering system}.
\newblock In \emph{2023 IEEE 17th International Conference on Semantic Computing (ICSC)}, pages 33--40.

\bibitem[{Aroca-Ouellette et~al.(2021)Aroca-Ouellette, Paik, Roncone, and Kann}]{arocaouellette2021prost}
Stéphane Aroca-Ouellette, Cory Paik, Alessandro Roncone, and Katharina Kann. 2021.
\newblock \href {http://arxiv.org/abs/2106.03634} {Prost: Physical reasoning of objects through space and time}.

\bibitem[{Bajaj et~al.(2018)Bajaj, Campos, Craswell, Deng, Gao, Liu, Majumder, McNamara, Mitra, Nguyen, Rosenberg, Song, Stoica, Tiwary, and Wang}]{msmacro}
Payal Bajaj, Daniel Campos, Nick Craswell, Li~Deng, Jianfeng Gao, Xiaodong Liu, Rangan Majumder, Andrew McNamara, Bhaskar Mitra, Tri Nguyen, Mir Rosenberg, Xia Song, Alina Stoica, Saurabh Tiwary, and Tong Wang. 2018.
\newblock \href {http://arxiv.org/abs/1611.09268} {Ms marco: A human generated machine reading comprehension dataset}.

\bibitem[{Bao et~al.(2021)Bao, Wang, and Zhao}]{bao-etal-2021-defending}
Rongzhou Bao, Jiayi Wang, and Hai Zhao. 2021.
\newblock \href {https://doi.org/10.18653/v1/2021.findings-acl.287} {Defending pre-trained language models from adversarial word substitution without performance sacrifice}.
\newblock In \emph{Findings of the Association for Computational Linguistics: ACL-IJCNLP 2021}, pages 3248--3258, Online. Association for Computational Linguistics.

\bibitem[{Bartolo et~al.(2020)Bartolo, Roberts, Welbl, Riedel, and Stenetorp}]{bartolo-etal-2020-beat}
Max Bartolo, Alastair Roberts, Johannes Welbl, Sebastian Riedel, and Pontus Stenetorp. 2020.
\newblock \href {https://doi.org/10.1162/tacl_a_00338} {Beat the {AI}: Investigating adversarial human annotation for reading comprehension}.
\newblock \emph{Transactions of the Association for Computational Linguistics}, 8:662--678.

\bibitem[{Beltagy et~al.(2020)Beltagy, Peters, and Cohan}]{longformer}
Iz~Beltagy, Matthew~E. Peters, and Arman Cohan. 2020.
\newblock \href {https://doi.org/10.48550/ARXIV.2004.05150} {Longformer: The long-document transformer}.

\bibitem[{Brown et~al.(2020)Brown, Mann, Ryder, Subbiah, Kaplan, Dhariwal, Neelakantan, Shyam, Sastry, Askell, Agarwal, Herbert-Voss, Krueger, Henighan, Child, Ramesh, Ziegler, Wu, Winter, Hesse, Chen, Sigler, Litwin, Gray, Chess, Clark, Berner, McCandlish, Radford, Sutskever, and Amodei}]{few_shot}
Tom~B. Brown, Benjamin Mann, Nick Ryder, Melanie Subbiah, Jared Kaplan, Prafulla Dhariwal, Arvind Neelakantan, Pranav Shyam, Girish Sastry, Amanda Askell, Sandhini Agarwal, Ariel Herbert-Voss, Gretchen Krueger, Tom Henighan, Rewon Child, Aditya Ramesh, Daniel~M. Ziegler, Jeffrey Wu, Clemens Winter, Christopher Hesse, Mark Chen, Eric Sigler, Mateusz Litwin, Scott Gray, Benjamin Chess, Jack Clark, Christopher Berner, Sam McCandlish, Alec Radford, Ilya Sutskever, and Dario Amodei. 2020.
\newblock \href {https://doi.org/10.48550/ARXIV.2005.14165} {Language models are few-shot learners}.

\bibitem[{Cao et~al.(2022)Cao, Li, Fang, Zhou, Gao, Zhan, and Tao}]{cao2022tasa}
Yu~Cao, Dianqi Li, Meng Fang, Tianyi Zhou, Jun Gao, Yibing Zhan, and Dacheng Tao. 2022.
\newblock \href {http://arxiv.org/abs/2210.15221} {Tasa: Deceiving question answering models by twin answer sentences attack}.

\bibitem[{Chen et~al.(2022)Chen, Zhang, and Choi}]{chen-etal-2022-rich}
Hung-Ting Chen, Michael Zhang, and Eunsol Choi. 2022.
\newblock \href {https://aclanthology.org/2022.emnlp-main.146} {Rich knowledge sources bring complex knowledge conflicts: Recalibrating models to reflect conflicting evidence}.
\newblock In \emph{Proceedings of the 2022 Conference on Empirical Methods in Natural Language Processing}, pages 2292--2307, Abu Dhabi, United Arab Emirates. Association for Computational Linguistics.

\bibitem[{Choi et~al.(2018)Choi, He, Iyyer, Yatskar, Yih, Choi, Liang, and Zettlemoyer}]{choi-etal-2018-quac}
Eunsol Choi, He~He, Mohit Iyyer, Mark Yatskar, Wen-tau Yih, Yejin Choi, Percy Liang, and Luke Zettlemoyer. 2018.
\newblock \href {https://doi.org/10.18653/v1/D18-1241} {{Q}u{AC}: Question answering in context}.
\newblock In \emph{Proceedings of the 2018 Conference on Empirical Methods in Natural Language Processing}, pages 2174--2184, Brussels, Belgium. Association for Computational Linguistics.

\bibitem[{Chowdhery et~al.(2022)Chowdhery, Narang, Devlin, Bosma, Mishra, Roberts, Barham, Chung, Sutton, Gehrmann, Schuh, Shi, Tsvyashchenko, Maynez, Rao, Barnes, Tay, Shazeer, Prabhakaran, Reif, Du, Hutchinson, Pope, Bradbury, Austin, Isard, Gur-Ari, Yin, Duke, Levskaya, Ghemawat, Dev, Michalewski, Garcia, Misra, Robinson, Fedus, Zhou, Ippolito, Luan, Lim, Zoph, Spiridonov, Sepassi, Dohan, Agrawal, Omernick, Dai, Pillai, Pellat, Lewkowycz, Moreira, Child, Polozov, Lee, Zhou, Wang, Saeta, Diaz, Firat, Catasta, Wei, Meier-Hellstern, Eck, Dean, Petrov, and Fiedel}]{palm}
Aakanksha Chowdhery, Sharan Narang, Jacob Devlin, Maarten Bosma, Gaurav Mishra, Adam Roberts, Paul Barham, Hyung~Won Chung, Charles Sutton, Sebastian Gehrmann, Parker Schuh, Kensen Shi, Sasha Tsvyashchenko, Joshua Maynez, Abhishek Rao, Parker Barnes, Yi~Tay, Noam Shazeer, Vinodkumar Prabhakaran, Emily Reif, Nan Du, Ben Hutchinson, Reiner Pope, James Bradbury, Jacob Austin, Michael Isard, Guy Gur-Ari, Pengcheng Yin, Toju Duke, Anselm Levskaya, Sanjay Ghemawat, Sunipa Dev, Henryk Michalewski, Xavier Garcia, Vedant Misra, Kevin Robinson, Liam Fedus, Denny Zhou, Daphne Ippolito, David Luan, Hyeontaek Lim, Barret Zoph, Alexander Spiridonov, Ryan Sepassi, David Dohan, Shivani Agrawal, Mark Omernick, Andrew~M. Dai, Thanumalayan~Sankaranarayana Pillai, Marie Pellat, Aitor Lewkowycz, Erica Moreira, Rewon Child, Oleksandr Polozov, Katherine Lee, Zongwei Zhou, Xuezhi Wang, Brennan Saeta, Mark Diaz, Orhan Firat, Michele Catasta, Jason Wei, Kathy Meier-Hellstern, Douglas Eck, Jeff Dean, Slav Petrov, and Noah Fiedel. 2022.
\newblock \href {https://doi.org/10.48550/ARXIV.2204.02311} {Palm: Scaling language modeling with pathways}.

\bibitem[{Cohen et~al.(2023)Cohen, Geva, Berant, and Globerson}]{crawling}
Roi Cohen, Mor Geva, Jonathan Berant, and Amir Globerson. 2023.
\newblock \href {https://doi.org/10.48550/ARXIV.2301.12810} {Crawling the internal knowledge-base of language models}.

\bibitem[{De~Cao et~al.(2021{\natexlab{a}})De~Cao, Aziz, and Titov}]{edit}
Nicola De~Cao, Wilker Aziz, and Ivan Titov. 2021{\natexlab{a}}.
\newblock \href {https://doi.org/10.48550/ARXIV.2104.08164} {Editing factual knowledge in language models}.

\bibitem[{De~Cao et~al.(2021{\natexlab{b}})De~Cao, Aziz, and Titov}]{editing}
Nicola De~Cao, Wilker Aziz, and Ivan Titov. 2021{\natexlab{b}}.
\newblock \href {https://doi.org/10.48550/ARXIV.2104.08164} {Editing factual knowledge in language models}.

\bibitem[{Devlin et~al.(2018)Devlin, Chang, Lee, and Toutanova}]{bert}
Jacob Devlin, Ming-Wei Chang, Kenton Lee, and Kristina Toutanova. 2018.
\newblock \href {https://doi.org/10.48550/ARXIV.1810.04805} {Bert: Pre-training of deep bidirectional transformers for language understanding}.

\bibitem[{Dhingra et~al.(2022)Dhingra, Cole, Eisenschlos, Gillick, Eisenstein, and Cohen}]{dhingra-etal-2022-time}
Bhuwan Dhingra, Jeremy~R. Cole, Julian~Martin Eisenschlos, Daniel Gillick, Jacob Eisenstein, and William~W. Cohen. 2022.
\newblock \href {https://doi.org/10.1162/tacl_a_00459} {Time-aware language models as temporal knowledge bases}.
\newblock \emph{Transactions of the Association for Computational Linguistics}, 10:257--273.

\bibitem[{Dzendzik et~al.(2021)Dzendzik, Foster, and Vogel}]{dzendzik-etal-2021-english}
Daria Dzendzik, Jennifer Foster, and Carl Vogel. 2021.
\newblock \href {https://doi.org/10.18653/v1/2021.emnlp-main.693} {{E}nglish machine reading comprehension datasets: A survey}.
\newblock In \emph{Proceedings of the 2021 Conference on Empirical Methods in Natural Language Processing}, pages 8784--8804, Online and Punta Cana, Dominican Republic. Association for Computational Linguistics.

\bibitem[{Eisenstein et~al.(2022)Eisenstein, Andor, Bohnet, Collins, and Mimno}]{eisenstein2022honest}
Jacob Eisenstein, Daniel Andor, Bernd Bohnet, Michael Collins, and David Mimno. 2022.
\newblock \href {http://arxiv.org/abs/2210.02498} {Honest students from untrusted teachers: Learning an interpretable question-answering pipeline from a pretrained language model}.

\bibitem[{Fang et~al.(2023)Fang, Wang, Zhou, Zhang, Song, and Chen}]{fang2023getting}
Tianqing Fang, Zhaowei Wang, Wenxuan Zhou, Hongming Zhang, Yangqiu Song, and Muhao Chen. 2023.
\newblock \href {http://arxiv.org/abs/2305.14970} {Getting sick after seeing a doctor? diagnosing and mitigating knowledge conflicts in event temporal reasoning}.

\bibitem[{Gan and Ng(2019)}]{gan-ng-2019-improving}
Wee~Chung Gan and Hwee~Tou Ng. 2019.
\newblock \href {https://doi.org/10.18653/v1/P19-1610} {Improving the robustness of question answering systems to question paraphrasing}.
\newblock In \emph{Proceedings of the 57th Annual Meeting of the Association for Computational Linguistics}, pages 6065--6075, Florence, Italy. Association for Computational Linguistics.

\bibitem[{Gardner et~al.(2020)Gardner, Artzi, Basmov, Berant, Bogin, Chen, Dasigi, Dua, Elazar, Gottumukkala, Gupta, Hajishirzi, Ilharco, Khashabi, Lin, Liu, Liu, Mulcaire, Ning, Singh, Smith, Subramanian, Tsarfaty, Wallace, Zhang, and Zhou}]{gardner-etal-2020-evaluating}
Matt Gardner, Yoav Artzi, Victoria Basmov, Jonathan Berant, Ben Bogin, Sihao Chen, Pradeep Dasigi, Dheeru Dua, Yanai Elazar, Ananth Gottumukkala, Nitish Gupta, Hannaneh Hajishirzi, Gabriel Ilharco, Daniel Khashabi, Kevin Lin, Jiangming Liu, Nelson~F. Liu, Phoebe Mulcaire, Qiang Ning, Sameer Singh, Noah~A. Smith, Sanjay Subramanian, Reut Tsarfaty, Eric Wallace, Ally Zhang, and Ben Zhou. 2020.
\newblock \href {https://doi.org/10.18653/v1/2020.findings-emnlp.117} {Evaluating models{'} local decision boundaries via contrast sets}.
\newblock In \emph{Findings of the Association for Computational Linguistics: EMNLP 2020}, pages 1307--1323, Online. Association for Computational Linguistics.

\bibitem[{Geva et~al.(2020)Geva, Schuster, Berant, and Levy}]{geva}
Mor Geva, Roei Schuster, Jonathan Berant, and Omer Levy. 2020.
\newblock \href {https://doi.org/10.48550/ARXIV.2012.14913} {Transformer feed-forward layers are key-value memories}.

\bibitem[{Gubelmann and Handschuh(2022)}]{gubelmann-handschuh-2022-context}
Reto Gubelmann and Siegfried Handschuh. 2022.
\newblock \href {https://doi.org/10.18653/v1/2022.acl-long.315} {Context matters: A pragmatic study of {PLM}s{'} negation understanding}.
\newblock In \emph{Proceedings of the 60th Annual Meeting of the Association for Computational Linguistics (Volume 1: Long Papers)}, pages 4602--4621, Dublin, Ireland. Association for Computational Linguistics.

\bibitem[{Hong et~al.(2023)Hong, Kim, Kang, Myaeng, and Whang}]{hong2023discern}
Giwon Hong, Jeonghwan Kim, Junmo Kang, Sung-Hyon Myaeng, and Joyce~Jiyoung Whang. 2023.
\newblock \href {http://arxiv.org/abs/2305.01579} {Discern and answer: Mitigating the impact of misinformation in retrieval-augmented models with discriminators}.

\bibitem[{Huang et~al.(2019)Huang, Bras, Bhagavatula, and Choi}]{cosmos}
Lifu Huang, Ronan~Le Bras, Chandra Bhagavatula, and Yejin Choi. 2019.
\newblock \href {https://doi.org/10.48550/ARXIV.1909.00277} {Cosmos qa: Machine reading comprehension with contextual commonsense reasoning}.

\bibitem[{Jia and Liang(2017)}]{jia-liang-2017-adversarial}
Robin Jia and Percy Liang. 2017.
\newblock \href {https://doi.org/10.18653/v1/D17-1215} {Adversarial examples for evaluating reading comprehension systems}.
\newblock In \emph{Proceedings of the 2017 Conference on Empirical Methods in Natural Language Processing}, pages 2021--2031, Copenhagen, Denmark. Association for Computational Linguistics.

\bibitem[{Jiang et~al.(2020{\natexlab{a}})Jiang, Xu, Araki, and Neubig}]{jiang2020know}
Zhengbao Jiang, Frank~F. Xu, Jun Araki, and Graham Neubig. 2020{\natexlab{a}}.
\newblock \href {http://arxiv.org/abs/1911.12543} {How can we know what language models know?}

\bibitem[{Jiang et~al.(2020{\natexlab{b}})Jiang, Xu, Araki, and Neubig}]{jiang-etal-2020-know}
Zhengbao Jiang, Frank~F. Xu, Jun Araki, and Graham Neubig. 2020{\natexlab{b}}.
\newblock \href {https://doi.org/10.1162/tacl_a_00324} {How can we know what language models know?}
\newblock \emph{Transactions of the Association for Computational Linguistics}, 8:423--438.

\bibitem[{Kamalloo et~al.(2023)Kamalloo, Dziri, Clarke, and Rafiei}]{kamalloo-etal-2023-evaluating}
Ehsan Kamalloo, Nouha Dziri, Charles Clarke, and Davood Rafiei. 2023.
\newblock \href {https://doi.org/10.18653/v1/2023.acl-long.307} {Evaluating open-domain question answering in the era of large language models}.
\newblock In \emph{Proceedings of the 61st Annual Meeting of the Association for Computational Linguistics (Volume 1: Long Papers)}, pages 5591--5606, Toronto, Canada. Association for Computational Linguistics.

\bibitem[{Kassner and Sch{\"u}tze(2020)}]{kassner-schutze-2020-negated}
Nora Kassner and Hinrich Sch{\"u}tze. 2020.
\newblock \href {https://doi.org/10.18653/v1/2020.acl-main.698} {Negated and misprimed probes for pretrained language models: Birds can talk, but cannot fly}.
\newblock In \emph{Proceedings of the 58th Annual Meeting of the Association for Computational Linguistics}, pages 7811--7818, Online. Association for Computational Linguistics.

\bibitem[{Kassner et~al.(2021)Kassner, Tafjord, Sch{\"u}tze, and Clark}]{kassner-etal-2021-beliefbank}
Nora Kassner, Oyvind Tafjord, Hinrich Sch{\"u}tze, and Peter Clark. 2021.
\newblock \href {https://doi.org/10.18653/v1/2021.emnlp-main.697} {{B}elief{B}ank: Adding memory to a pre-trained language model for a systematic notion of belief}.
\newblock In \emph{Proceedings of the 2021 Conference on Empirical Methods in Natural Language Processing}, pages 8849--8861, Online and Punta Cana, Dominican Republic. Association for Computational Linguistics.

\bibitem[{Kaushik and Lipton(2018{\natexlab{a}})}]{how_much}
Divyansh Kaushik and Zachary~C. Lipton. 2018{\natexlab{a}}.
\newblock \href {https://doi.org/10.48550/ARXIV.1808.04926} {How much reading does reading comprehension require? a critical investigation of popular benchmarks}.

\bibitem[{Kaushik and Lipton(2018{\natexlab{b}})}]{kaushik-lipton-2018-much}
Divyansh Kaushik and Zachary~C. Lipton. 2018{\natexlab{b}}.
\newblock \href {https://doi.org/10.18653/v1/D18-1546} {How much reading does reading comprehension require? a critical investigation of popular benchmarks}.
\newblock In \emph{Proceedings of the 2018 Conference on Empirical Methods in Natural Language Processing}, pages 5010--5015, Brussels, Belgium. Association for Computational Linguistics.

\bibitem[{Khashabi et~al.(2020)Khashabi, Khot, and Sabharwal}]{khashabi-etal-2020-bang}
Daniel Khashabi, Tushar Khot, and Ashish Sabharwal. 2020.
\newblock \href {https://doi.org/10.18653/v1/2020.emnlp-main.12} {More bang for your buck: Natural perturbation for robust question answering}.
\newblock In \emph{Proceedings of the 2020 Conference on Empirical Methods in Natural Language Processing (EMNLP)}, pages 163--170, Online. Association for Computational Linguistics.

\bibitem[{Kwiatkowski et~al.(2019)Kwiatkowski, Palomaki, Redfield, Collins, Parikh, Alberti, Epstein, Polosukhin, Devlin, Lee, Toutanova, Jones, Kelcey, Chang, Dai, Uszkoreit, Le, and Petrov}]{kwiatkowski-etal-2019-natural}
Tom Kwiatkowski, Jennimaria Palomaki, Olivia Redfield, Michael Collins, Ankur Parikh, Chris Alberti, Danielle Epstein, Illia Polosukhin, Jacob Devlin, Kenton Lee, Kristina Toutanova, Llion Jones, Matthew Kelcey, Ming-Wei Chang, Andrew~M. Dai, Jakob Uszkoreit, Quoc Le, and Slav Petrov. 2019.
\newblock \href {https://doi.org/10.1162/tacl_a_00276} {Natural questions: A benchmark for question answering research}.
\newblock \emph{Transactions of the Association for Computational Linguistics}, 7:452--466.

\bibitem[{Lai et~al.(2017)Lai, Xie, Liu, Yang, and Hovy}]{race}
Guokun Lai, Qizhe Xie, Hanxiao Liu, Yiming Yang, and Eduard Hovy. 2017.
\newblock \href {https://doi.org/10.48550/ARXIV.1704.04683} {Race: Large-scale reading comprehension dataset from examinations}.

\bibitem[{Lan et~al.(2020)Lan, Chen, Goodman, Gimpel, Sharma, and Soricut}]{lan2020albert}
Zhenzhong Lan, Mingda Chen, Sebastian Goodman, Kevin Gimpel, Piyush Sharma, and Radu Soricut. 2020.
\newblock \href {http://arxiv.org/abs/1909.11942} {Albert: A lite bert for self-supervised learning of language representations}.

\bibitem[{Lewis et~al.(2019)Lewis, Liu, Goyal, Ghazvininejad, Mohamed, Levy, Stoyanov, and Zettlemoyer}]{BART}
Mike Lewis, Yinhan Liu, Naman Goyal, Marjan Ghazvininejad, Abdelrahman Mohamed, Omer Levy, Ves Stoyanov, and Luke Zettlemoyer. 2019.
\newblock \href {https://doi.org/10.48550/ARXIV.1910.13461} {Bart: Denoising sequence-to-sequence pre-training for natural language generation, translation, and comprehension}.

\bibitem[{Li et~al.(2022)Li, Rawat, Zaheer, Wang, Lukasik, Veit, Yu, and Kumar}]{control}
Daliang Li, Ankit~Singh Rawat, Manzil Zaheer, Xin Wang, Michal Lukasik, Andreas Veit, Felix Yu, and Sanjiv Kumar. 2022.
\newblock \href {https://doi.org/10.48550/ARXIV.2211.05110} {Large language models with controllable working memory}.

\bibitem[{Li et~al.(2020)Li, Ma, Guo, Xue, and Qiu}]{li-etal-2020-bert-attack}
Linyang Li, Ruotian Ma, Qipeng Guo, Xiangyang Xue, and Xipeng Qiu. 2020.
\newblock \href {https://doi.org/10.18653/v1/2020.emnlp-main.500} {{BERT}-{ATTACK}: Adversarial attack against {BERT} using {BERT}}.
\newblock In \emph{Proceedings of the 2020 Conference on Empirical Methods in Natural Language Processing (EMNLP)}, pages 6193--6202, Online. Association for Computational Linguistics.

\bibitem[{Liang et~al.(2022)Liang, Bommasani, Lee, Tsipras, Soylu, Yasunaga, Zhang, Narayanan, Wu, Kumar, Newman, Yuan, Yan, Zhang, Cosgrove, Manning, Ré, Acosta-Navas, Hudson, Zelikman, Durmus, Ladhak, Rong, Ren, Yao, Wang, Santhanam, Orr, Zheng, Yuksekgonul, Suzgun, Kim, Guha, Chatterji, Khattab, Henderson, Huang, Chi, Xie, Santurkar, Ganguli, Hashimoto, Icard, Zhang, Chaudhary, Wang, Li, Mai, Zhang, and Koreeda}]{liang2022holistic}
Percy Liang, Rishi Bommasani, Tony Lee, Dimitris Tsipras, Dilara Soylu, Michihiro Yasunaga, Yian Zhang, Deepak Narayanan, Yuhuai Wu, Ananya Kumar, Benjamin Newman, Binhang Yuan, Bobby Yan, Ce~Zhang, Christian Cosgrove, Christopher~D. Manning, Christopher Ré, Diana Acosta-Navas, Drew~A. Hudson, Eric Zelikman, Esin Durmus, Faisal Ladhak, Frieda Rong, Hongyu Ren, Huaxiu Yao, Jue Wang, Keshav Santhanam, Laurel Orr, Lucia Zheng, Mert Yuksekgonul, Mirac Suzgun, Nathan Kim, Neel Guha, Niladri Chatterji, Omar Khattab, Peter Henderson, Qian Huang, Ryan Chi, Sang~Michael Xie, Shibani Santurkar, Surya Ganguli, Tatsunori Hashimoto, Thomas Icard, Tianyi Zhang, Vishrav Chaudhary, William Wang, Xuechen Li, Yifan Mai, Yuhui Zhang, and Yuta Koreeda. 2022.
\newblock \href {http://arxiv.org/abs/2211.09110} {Holistic evaluation of language models}.

\bibitem[{Liu et~al.(2019)Liu, Ott, Goyal, Du, Joshi, Chen, Levy, Lewis, Zettlemoyer, and Stoyanov}]{liu2019roberta}
Yinhan Liu, Myle Ott, Naman Goyal, Jingfei Du, Mandar Joshi, Danqi Chen, Omer Levy, Mike Lewis, Luke Zettlemoyer, and Veselin Stoyanov. 2019.
\newblock \href {http://arxiv.org/abs/1907.11692} {Roberta: A robustly optimized bert pretraining approach}.

\bibitem[{Longpre et~al.(2021)Longpre, Perisetla, Chen, Ramesh, DuBois, and Singh}]{longpre-etal-2021-entity}
Shayne Longpre, Kartik Perisetla, Anthony Chen, Nikhil Ramesh, Chris DuBois, and Sameer Singh. 2021.
\newblock \href {https://doi.org/10.18653/v1/2021.emnlp-main.565} {Entity-based knowledge conflicts in question answering}.
\newblock In \emph{Proceedings of the 2021 Conference on Empirical Methods in Natural Language Processing}, pages 7052--7063, Online and Punta Cana, Dominican Republic. Association for Computational Linguistics.

\bibitem[{Maharana and Bansal(2020)}]{maharana-bansal-2020-adversarial}
Adyasha Maharana and Mohit Bansal. 2020.
\newblock \href {https://doi.org/10.18653/v1/2020.findings-emnlp.333} {Adversarial augmentation policy search for domain and cross-lingual generalization in reading comprehension}.
\newblock In \emph{Findings of the Association for Computational Linguistics: EMNLP 2020}, pages 3723--3738, Online. Association for Computational Linguistics.

\bibitem[{Meng et~al.(2022)Meng, Bau, Andonian, and Belinkov}]{gpt_edit}
Kevin Meng, David Bau, Alex Andonian, and Yonatan Belinkov. 2022.
\newblock \href {https://doi.org/10.48550/ARXIV.2202.05262} {Locating and editing factual associations in gpt}.

\bibitem[{Michel et~al.(2019)Michel, Li, Neubig, and Pino}]{michel-etal-2019-evaluation}
Paul Michel, Xian Li, Graham Neubig, and Juan Pino. 2019.
\newblock \href {https://doi.org/10.18653/v1/N19-1314} {On evaluation of adversarial perturbations for sequence-to-sequence models}.
\newblock In \emph{Proceedings of the 2019 Conference of the North {A}merican Chapter of the Association for Computational Linguistics: Human Language Technologies, Volume 1 (Long and Short Papers)}, pages 3103--3114, Minneapolis, Minnesota. Association for Computational Linguistics.

\bibitem[{Mihaylov et~al.(2018)Mihaylov, Clark, Khot, and Sabharwal}]{openbookqa}
Todor Mihaylov, Peter Clark, Tushar Khot, and Ashish Sabharwal. 2018.
\newblock \href {https://doi.org/10.48550/ARXIV.1809.02789} {Can a suit of armor conduct electricity? a new dataset for open book question answering}.

\bibitem[{Mitchell et~al.(2021)Mitchell, Lin, Bosselut, Finn, and Manning}]{fast_edit}
Eric Mitchell, Charles Lin, Antoine Bosselut, Chelsea Finn, and Christopher~D. Manning. 2021.
\newblock \href {https://doi.org/10.48550/ARXIV.2110.11309} {Fast model editing at scale}.

\bibitem[{Moon and Fan(2020)}]{moon-fan-2020-ask}
Sungrim~(Riea) Moon and Jungwei Fan. 2020.
\newblock \href {https://doi.org/10.18653/v1/2020.clinicalnlp-1.13} {How you ask matters: The effect of paraphrastic questions to {BERT} performance on a clinical {SQ}u{AD} dataset}.
\newblock In \emph{Proceedings of the 3rd Clinical Natural Language Processing Workshop}, pages 111--116, Online. Association for Computational Linguistics.

\bibitem[{Mudrakarta et~al.(2018)Mudrakarta, Taly, Sundararajan, and Dhamdhere}]{ignore}
Pramod~Kaushik Mudrakarta, Ankur Taly, Mukund Sundararajan, and Kedar Dhamdhere. 2018.
\newblock \href {https://doi.org/10.48550/ARXIV.1805.05492} {Did the model understand the question?}

\bibitem[{Neeman et~al.(2022)Neeman, Aharoni, Honovich, Choshen, Szpektor, and Abend}]{dist}
Ella Neeman, Roee Aharoni, Or~Honovich, Leshem Choshen, Idan Szpektor, and Omri Abend. 2022.
\newblock \href {https://doi.org/10.48550/ARXIV.2211.05655} {Disentqa: Disentangling parametric and contextual knowledge with counterfactual question answering}.

\bibitem[{Onoe et~al.(2022)Onoe, Zhang, Choi, and Durrett}]{onoe-etal-2022-entity}
Yasumasa Onoe, Michael Zhang, Eunsol Choi, and Greg Durrett. 2022.
\newblock \href {https://doi.org/10.18653/v1/2022.findings-naacl.52} {Entity cloze by date: What {LM}s know about unseen entities}.
\newblock In \emph{Findings of the Association for Computational Linguistics: NAACL 2022}, pages 693--702, Seattle, United States. Association for Computational Linguistics.

\bibitem[{OpenAI(2023{\natexlab{a}})}]{openai_2023}
OpenAI. 2023{\natexlab{a}}.
\newblock \href {https://openai.com/blog/chatgpt/} {Chatgpt: Optimizing language models for dialogue}.

\bibitem[{OpenAI(2023{\natexlab{b}})}]{openai2023gpt4}
OpenAI. 2023{\natexlab{b}}.
\newblock \href {http://arxiv.org/abs/2303.08774} {Gpt-4 technical report}.

\bibitem[{Pal et~al.(2022)Pal, Umapathi, and Sankarasubbu}]{pal2022medmcqa}
Ankit Pal, Logesh~Kumar Umapathi, and Malaikannan Sankarasubbu. 2022.
\newblock \href {http://arxiv.org/abs/2203.14371} {Medmcqa : A large-scale multi-subject multi-choice dataset for medical domain question answering}.

\bibitem[{Pan et~al.(2021)Pan, Chen, Kan, and Wang}]{contra}
Liangming Pan, Wenhu Chen, Min-Yen Kan, and William~Yang Wang. 2021.
\newblock \href {https://doi.org/10.48550/ARXIV.2110.07803} {Contraqa: Question answering under contradicting contexts}.

\bibitem[{Pan et~al.(2023)Pan, Pan, Chen, Nakov, Kan, and Wang}]{pan2023risk}
Yikang Pan, Liangming Pan, Wenhu Chen, Preslav Nakov, Min-Yen Kan, and William~Yang Wang. 2023.
\newblock \href {http://arxiv.org/abs/2305.13661} {On the risk of misinformation pollution with large language models}.

\bibitem[{Petroni et~al.(2019)Petroni, Rocktäschel, Lewis, Bakhtin, Wu, Miller, and Riedel}]{petroni}
Fabio Petroni, Tim Rocktäschel, Patrick Lewis, Anton Bakhtin, Yuxiang Wu, Alexander~H. Miller, and Sebastian Riedel. 2019.
\newblock \href {https://doi.org/10.48550/ARXIV.1909.01066} {Language models as knowledge bases?}

\bibitem[{Qian et~al.(2022)Qian, Ross, Fernandes, Smith, Kiela, and Williams}]{qian-etal-2022-perturbation}
Rebecca Qian, Candace Ross, Jude Fernandes, Eric~Michael Smith, Douwe Kiela, and Adina Williams. 2022.
\newblock \href {https://aclanthology.org/2022.emnlp-main.646} {Perturbation augmentation for fairer {NLP}}.
\newblock In \emph{Proceedings of the 2022 Conference on Empirical Methods in Natural Language Processing}, pages 9496--9521, Abu Dhabi, United Arab Emirates. Association for Computational Linguistics.

\bibitem[{Radford et~al.(2018)Radford, Narasimhan, Salimans, and Sutskever}]{radford2018improving}
Alec Radford, Karthik Narasimhan, Tim Salimans, and Ilya Sutskever. 2018.
\newblock Improving language understanding by generative pre-training.

\bibitem[{Radford et~al.(2019)Radford, Wu, Child, Luan, Amodei, and Sutskever}]{radford_language_2019}
Alec Radford, Jeff Wu, Rewon Child, D.~Luan, Dario Amodei, and Ilya Sutskever. 2019.
\newblock \href {https://www.semanticscholar.org/paper/Language-Models-are-Unsupervised-Multitask-Learners-Radford-Wu/9405cc0d6169988371b2755e573cc28650d14dfe} {Language {Models} are {Unsupervised} {Multitask} {Learners}}.

\bibitem[{Raffel et~al.(2019)Raffel, Shazeer, Roberts, Lee, Narang, Matena, Zhou, Li, and Liu}]{t5}
Colin Raffel, Noam Shazeer, Adam Roberts, Katherine Lee, Sharan Narang, Michael Matena, Yanqi Zhou, Wei Li, and Peter~J. Liu. 2019.
\newblock \href {https://doi.org/10.48550/ARXIV.1910.10683} {Exploring the limits of transfer learning with a unified text-to-text transformer}.

\bibitem[{Rajpurkar et~al.(2018)Rajpurkar, Jia, and Liang}]{rajpurkar-etal-2018-know}
Pranav Rajpurkar, Robin Jia, and Percy Liang. 2018.
\newblock \href {https://doi.org/10.18653/v1/P18-2124} {Know what you don{'}t know: Unanswerable questions for {SQ}u{AD}}.
\newblock In \emph{Proceedings of the 56th Annual Meeting of the Association for Computational Linguistics (Volume 2: Short Papers)}, pages 784--789, Melbourne, Australia. Association for Computational Linguistics.

\bibitem[{Rajpurkar et~al.(2016)Rajpurkar, Zhang, Lopyrev, and Liang}]{squad}
Pranav Rajpurkar, Jian Zhang, Konstantin Lopyrev, and Percy Liang. 2016.
\newblock \href {https://doi.org/10.48550/ARXIV.1606.05250} {Squad: 100,000+ questions for machine comprehension of text}.

\bibitem[{Reddy et~al.(2019)Reddy, Chen, and Manning}]{reddy-etal-2019-coqa}
Siva Reddy, Danqi Chen, and Christopher~D. Manning. 2019.
\newblock \href {https://doi.org/10.1162/tacl_a_00266} {{C}o{QA}: A conversational question answering challenge}.
\newblock \emph{Transactions of the Association for Computational Linguistics}, 7:249--266.

\bibitem[{Ribeiro et~al.(2018)Ribeiro, Singh, and Guestrin}]{ribeiro-etal-2018-semantically}
Marco~Tulio Ribeiro, Sameer Singh, and Carlos Guestrin. 2018.
\newblock \href {https://doi.org/10.18653/v1/P18-1079} {Semantically equivalent adversarial rules for debugging {NLP} models}.
\newblock In \emph{Proceedings of the 56th Annual Meeting of the Association for Computational Linguistics (Volume 1: Long Papers)}, pages 856--865, Melbourne, Australia. Association for Computational Linguistics.

\bibitem[{Roberts et~al.(2020)Roberts, Raffel, and Shazeer}]{roberts-etal-2020-much}
Adam Roberts, Colin Raffel, and Noam Shazeer. 2020.
\newblock \href {https://doi.org/10.18653/v1/2020.emnlp-main.437} {How much knowledge can you pack into the parameters of a language model?}
\newblock In \emph{Proceedings of the 2020 Conference on Empirical Methods in Natural Language Processing (EMNLP)}, pages 5418--5426, Online. Association for Computational Linguistics.

\bibitem[{Rogers et~al.(2020)Rogers, Kovaleva, Downey, and Rumshisky}]{quail}
Anna Rogers, Olga Kovaleva, Matthew Downey, and Anna Rumshisky. 2020.
\newblock Getting closer to ai complete question answering: A set of prerequisite real tasks.
\newblock In \emph{AAAI Conference on Artificial Intelligence}.

\bibitem[{Sanh et~al.(2020)Sanh, Debut, Chaumond, and Wolf}]{sanh2020distilbert}
Victor Sanh, Lysandre Debut, Julien Chaumond, and Thomas Wolf. 2020.
\newblock \href {http://arxiv.org/abs/1910.01108} {Distilbert, a distilled version of bert: smaller, faster, cheaper and lighter}.

\bibitem[{Schlegel et~al.(2021)Schlegel, Nenadic, and Batista-Navarro}]{schlegel2021semantics}
Viktor Schlegel, Goran Nenadic, and Riza Batista-Navarro. 2021.
\newblock \href {http://arxiv.org/abs/2012.04056} {Semantics altering modifications for evaluating comprehension in machine reading}.

\bibitem[{Sen and Saffari(2020)}]{sen-saffari-2020-models}
Priyanka Sen and Amir Saffari. 2020.
\newblock \href {https://doi.org/10.18653/v1/2020.emnlp-main.190} {What do models learn from question answering datasets?}
\newblock In \emph{Proceedings of the 2020 Conference on Empirical Methods in Natural Language Processing (EMNLP)}, pages 2429--2438, Online. Association for Computational Linguistics.

\bibitem[{Shaier et~al.(2023{\natexlab{a}})Shaier, Bennett, Hunter, and Kann}]{shaier-etal-2023-emerging}
Sagi Shaier, Kevin Bennett, Lawrence Hunter, and Katharina Kann. 2023{\natexlab{a}}.
\newblock \href {https://aclanthology.org/2023.ijcnlp-main.36} {Emerging challenges in personalized medicine: Assessing demographic effects on biomedical question answering systems}.
\newblock In \emph{Proceedings of the 13th International Joint Conference on Natural Language Processing and the 3rd Conference of the Asia-Pacific Chapter of the Association for Computational Linguistics (Volume 1: Long Papers)}, pages 540--550, Nusa Dua, Bali. Association for Computational Linguistics.

\bibitem[{Shaier et~al.(2024{\natexlab{a}})Shaier, Bennett, Hunter, and von~der Wense}]{shaier-etal-2024-comparing}
Sagi Shaier, Kevin Bennett, Lawrence Hunter, and Katharina von~der Wense. 2024{\natexlab{a}}.
\newblock \href {https://aclanthology.org/2024.eacl-long.46} {Comparing template-based and template-free language model probing}.
\newblock In \emph{Proceedings of the 18th Conference of the European Chapter of the Association for Computational Linguistics (Volume 1: Long Papers)}, pages 766--776, St. Julian{'}s, Malta. Association for Computational Linguistics.

\bibitem[{Shaier et~al.(2024{\natexlab{b}})Shaier, Hunter, and von~der Wense}]{shaier-etal-2024-say}
Sagi Shaier, Lawrence Hunter, and Katharina von~der Wense. 2024{\natexlab{b}}.
\newblock \href {https://aclanthology.org/2024.findings-acl.491} {It is not about what you say, it is about how you say it: A surprisingly simple approach for improving reading comprehension}.
\newblock In \emph{Findings of the Association for Computational Linguistics ACL 2024}, pages 8292--8305, Bangkok, Thailand and virtual meeting. Association for Computational Linguistics.

\bibitem[{Shaier et~al.(2023{\natexlab{b}})Shaier, Hunter, and von~der Wense}]{shaier2023stochasticparrotsimitatingtell}
Sagi Shaier, Lawrence~E. Hunter, and Katharina von~der Wense. 2023{\natexlab{b}}.
\newblock \href {http://arxiv.org/abs/2310.10583} {Who are all the stochastic parrots imitating? they should tell us!}

\bibitem[{Shaier et~al.(2024{\natexlab{c}})Shaier, Kobren, and Ogren}]{shaier2024adaptivequestionansweringenhancing}
Sagi Shaier, Ari Kobren, and Philip Ogren. 2024{\natexlab{c}}.
\newblock \href {http://arxiv.org/abs/2410.04241} {Adaptive question answering: Enhancing language model proficiency for addressing knowledge conflicts with source citations}.

\bibitem[{Shi et~al.(2023)Shi, Chen, Misra, Scales, Dohan, Chi, Schärli, and Zhou}]{shi2023large}
Freda Shi, Xinyun Chen, Kanishka Misra, Nathan Scales, David Dohan, Ed~Chi, Nathanael Schärli, and Denny Zhou. 2023.
\newblock \href {http://arxiv.org/abs/2302.00093} {Large language models can be easily distracted by irrelevant context}.

\bibitem[{Shin et~al.(2020)Shin, Razeghi, Logan~IV, Wallace, and Singh}]{shin-etal-2020-autoprompt}
Taylor Shin, Yasaman Razeghi, Robert~L. Logan~IV, Eric Wallace, and Sameer Singh. 2020.
\newblock \href {https://doi.org/10.18653/v1/2020.emnlp-main.346} {{A}uto{P}rompt: {E}liciting {K}nowledge from {L}anguage {M}odels with {A}utomatically {G}enerated {P}rompts}.
\newblock In \emph{Proceedings of the 2020 Conference on Empirical Methods in Natural Language Processing (EMNLP)}, pages 4222--4235, Online. Association for Computational Linguistics.

\bibitem[{Si et~al.(2023)Si, Gan, Yang, Wang, Wang, Boyd-Graber, and Wang}]{si2023prompting}
Chenglei Si, Zhe Gan, Zhengyuan Yang, Shuohang Wang, Jianfeng Wang, Jordan Boyd-Graber, and Lijuan Wang. 2023.
\newblock \href {http://arxiv.org/abs/2210.09150} {Prompting gpt-3 to be reliable}.

\bibitem[{Si et~al.(2019)Si, Wang, Kan, and Jiang}]{si2019does}
Chenglei Si, Shuohang Wang, Min-Yen Kan, and Jing Jiang. 2019.
\newblock \href {http://arxiv.org/abs/1910.12391} {What does bert learn from multiple-choice reading comprehension datasets?}

\bibitem[{Si et~al.(2021)Si, Yang, Cui, Ma, Liu, and Wang}]{si-etal-2021-benchmarking}
Chenglei Si, Ziqing Yang, Yiming Cui, Wentao Ma, Ting Liu, and Shijin Wang. 2021.
\newblock \href {https://doi.org/10.18653/v1/2021.findings-acl.56} {Benchmarking robustness of machine reading comprehension models}.
\newblock In \emph{Findings of the Association for Computational Linguistics: ACL-IJCNLP 2021}, pages 634--644, Online. Association for Computational Linguistics.

\bibitem[{Sung et~al.(2021)Sung, Lee, Yi, Jeon, Kim, and Kang}]{sung-etal-2021-language}
Mujeen Sung, Jinhyuk Lee, Sean Yi, Minji Jeon, Sungdong Kim, and Jaewoo Kang. 2021.
\newblock \href {https://doi.org/10.18653/v1/2021.emnlp-main.388} {Can language models be biomedical knowledge bases?}
\newblock In \emph{Proceedings of the 2021 Conference on Empirical Methods in Natural Language Processing}, pages 4723--4734, Online and Punta Cana, Dominican Republic. Association for Computational Linguistics.

\bibitem[{Varshney et~al.(2023)Varshney, Parmar, Patel, Handa, Sarkar, Luo, and Baral}]{varshney2023nlp}
Neeraj Varshney, Mihir Parmar, Nisarg Patel, Divij Handa, Sayantan Sarkar, Man Luo, and Chitta Baral. 2023.
\newblock \href {http://arxiv.org/abs/2305.12096} {Can nlp models correctly reason over contexts that break the common assumptions?}

\bibitem[{Vaswani et~al.(2017)Vaswani, Shazeer, Parmar, Uszkoreit, Jones, Gomez, Kaiser, and Polosukhin}]{transformer}
Ashish Vaswani, Noam Shazeer, Niki Parmar, Jakob Uszkoreit, Llion Jones, Aidan~N Gomez, \L~ukasz Kaiser, and Illia Polosukhin. 2017.
\newblock \href {https://proceedings.neurips.cc/paper_files/paper/2017/file/3f5ee243547dee91fbd053c1c4a845aa-Paper.pdf} {Attention is all you need}.
\newblock In \emph{Advances in Neural Information Processing Systems}, volume~30. Curran Associates, Inc.

\bibitem[{Wang and Bansal(2018)}]{wang-bansal-2018-robust}
Yicheng Wang and Mohit Bansal. 2018.
\newblock \href {https://doi.org/10.18653/v1/N18-2091} {Robust machine comprehension models via adversarial training}.
\newblock In \emph{Proceedings of the 2018 Conference of the North {A}merican Chapter of the Association for Computational Linguistics: Human Language Technologies, Volume 2 (Short Papers)}, pages 575--581, New Orleans, Louisiana. Association for Computational Linguistics.

\bibitem[{Wei et~al.(2021)Wei, Bosma, Zhao, Guu, Yu, Lester, Du, Dai, and Le}]{flan}
Jason Wei, Maarten Bosma, Vincent~Y. Zhao, Kelvin Guu, Adams~Wei Yu, Brian Lester, Nan Du, Andrew~M. Dai, and Quoc~V. Le. 2021.
\newblock \href {https://doi.org/10.48550/ARXIV.2109.01652} {Finetuned language models are zero-shot learners}.

\bibitem[{Welbl et~al.(2017)Welbl, Liu, and Gardner}]{Welbl2017CrowdsourcingMC}
Johannes Welbl, Nelson~F. Liu, and Matt Gardner. 2017.
\newblock Crowdsourcing multiple choice science questions.
\newblock \emph{ArXiv}, abs/1707.06209.

\bibitem[{Xie et~al.(2023)Xie, Zhang, Chen, Lou, and Su}]{xie2023adaptive}
Jian Xie, Kai Zhang, Jiangjie Chen, Renze Lou, and Yu~Su. 2023.
\newblock \href {http://arxiv.org/abs/2305.13300} {Adaptive chameleon or stubborn sloth: Unraveling the behavior of large language models in knowledge clashes}.

\bibitem[{Yan et~al.(2022)Yan, Xiao, Mukherjee, Lin, Jia, and Ren}]{yan-etal-2022-robustness}
Jun Yan, Yang Xiao, Sagnik Mukherjee, Bill~Yuchen Lin, Robin Jia, and Xiang Ren. 2022.
\newblock \href {https://doi.org/10.18653/v1/2022.naacl-main.37} {On the robustness of reading comprehension models to entity renaming}.
\newblock In \emph{Proceedings of the 2022 Conference of the North American Chapter of the Association for Computational Linguistics: Human Language Technologies}, pages 508--520, Seattle, United States. Association for Computational Linguistics.

\bibitem[{Yang et~al.(2015)Yang, Yih, and Meek}]{yang-etal-2015-wikiqa}
Yi~Yang, Wen-tau Yih, and Christopher Meek. 2015.
\newblock \href {https://doi.org/10.18653/v1/D15-1237} {{W}iki{QA}: A challenge dataset for open-domain question answering}.
\newblock In \emph{Proceedings of the 2015 Conference on Empirical Methods in Natural Language Processing}, pages 2013--2018, Lisbon, Portugal. Association for Computational Linguistics.

\bibitem[{Yin et~al.(2023)Yin, Sun, Guo, Wu, Qiu, and Huang}]{know?}
Zhangyue Yin, Qiushi Sun, Qipeng Guo, Jiawen Wu, Xipeng Qiu, and Xuanjing Huang. 2023.
\newblock \href {http://arxiv.org/abs/2305.18153} {Do large language models know what they don't know?}

\bibitem[{Zaheer et~al.(2020)Zaheer, Guruganesh, Dubey, Ainslie, Alberti, Ontanon, Pham, Ravula, Wang, Yang, and Ahmed}]{bigbird}
Manzil Zaheer, Guru Guruganesh, Avinava Dubey, Joshua Ainslie, Chris Alberti, Santiago Ontanon, Philip Pham, Anirudh Ravula, Qifan Wang, Li~Yang, and Amr Ahmed. 2020.
\newblock \href {https://doi.org/10.48550/ARXIV.2007.14062} {Big bird: Transformers for longer sequences}.

\bibitem[{Zeng et~al.(2020)Zeng, Li, Li, Hu, and Hu}]{mcsurvey}
Changchang Zeng, Shaobo Li, Qin Li, Jie Hu, and Jianjun Hu. 2020.
\newblock \href {http://arxiv.org/abs/2006.11880} {A survey on machine reading comprehension: Tasks, evaluation metrics and benchmark datasets}.

\bibitem[{Zhong et~al.(2021)Zhong, Friedman, and Chen}]{zhong-etal-2021-factual}
Zexuan Zhong, Dan Friedman, and Danqi Chen. 2021.
\newblock \href {https://doi.org/10.18653/v1/2021.naacl-main.398} {Factual probing is [{MASK}]: Learning vs. learning to recall}.
\newblock In \emph{Proceedings of the 2021 Conference of the North American Chapter of the Association for Computational Linguistics: Human Language Technologies}, pages 5017--5033, Online. Association for Computational Linguistics.

\bibitem[{Zhong et~al.(2023)Zhong, Wu, Manning, Potts, and Chen}]{zhong2023mquake}
Zexuan Zhong, Zhengxuan Wu, Christopher~D. Manning, Christopher Potts, and Danqi Chen. 2023.
\newblock \href {http://arxiv.org/abs/2305.14795} {Mquake: Assessing knowledge editing in language models via multi-hop questions}.

\bibitem[{Zhou et~al.(2023)Zhou, Zhang, Poon, and Chen}]{zhou2023contextfaithful}
Wenxuan Zhou, Sheng Zhang, Hoifung Poon, and Muhao Chen. 2023.
\newblock \href {http://arxiv.org/abs/2303.11315} {Context-faithful prompting for large language models}.

\bibitem[{Zhu et~al.(2020)Zhu, Rawat, Zaheer, Bhojanapalli, Li, Yu, and Kumar}]{mod}
Chen Zhu, Ankit~Singh Rawat, Manzil Zaheer, Srinadh Bhojanapalli, Daliang Li, Felix Yu, and Sanjiv Kumar. 2020.
\newblock \href {https://doi.org/10.48550/ARXIV.2012.00363} {Modifying memories in transformer models}.

\bibitem[{Étienne Fortier-Dubois and Rosati(2023)}]{fortierdubois2023using}
Étienne Fortier-Dubois and Domenic Rosati. 2023.
\newblock \href {http://arxiv.org/abs/2211.05598} {Using contradictions improves question answering systems}.

\end{thebibliography}
